\documentclass[journal]{IEEEtran}
\ifCLASSINFOpdf
   \usepackage[pdftex]{graphicx}
   \graphicspath{{../pdf/}{../jpeg/}}
   \DeclareGraphicsExtensions{.pdf,.jpeg,.png}
\else
   \usepackage[dvips]{graphicx}
   \graphicspath{{../eps/}}
   \DeclareGraphicsExtensions{.eps}
\fi
\usepackage{booktabs}
\usepackage{subfigure}
\usepackage[cmex10]{amsmath}
\usepackage{color}
\usepackage{algorithmic}
\usepackage{algorithm}
\usepackage{multirow}
\usepackage[misc]{ifsym}
\usepackage{amsfonts}
\usepackage{url}
\usepackage{array}
\usepackage{footmisc}
\newcommand{\etal}{\emph{et al.\ }}
\newcommand{\tabincell}[2]{\begin{tabular}{@{}#1@{}}#2\end{tabular}}
\newcommand{\minitab}[2][l]{\begin{tabular}{#1}#2\end{tabular}}
\newcolumntype{C}[1]{>{\centering}p{#1}}
\newcolumntype{L}[1]{>{\raggedleft}p{#1}}
\newcolumntype{R}[1]{>{\raggedright}p{#1}}
\newcolumntype{I}{!{\vrule width 3pt}}
\newlength\savedwidth

\newlength\savewidth

\begin{document}
\title{DehazeNet: An End-to-End System for Single Image Haze Removal}

\author{Bolun~Cai,
        Xiangmin~Xu,~\IEEEmembership{Member,~IEEE,}
        Kui~Jia,~\IEEEmembership{Member,~IEEE,}
        Chunmei~Qing,~\IEEEmembership{Member,~IEEE,}
        and~Dacheng~Tao,~\IEEEmembership{Fellow,~IEEE}
\thanks{B. Cai, X. Xu (\Letter) and C. Qing are with
 School of Electronic and Information Engineering,
 South China University of Technology,
 Wushan RD., Tianhe District, Guangzhou, P.R.China.
 E-mail: \{caibolun@gmail.com, xmxu@scut.edu.cn, qchm@scut.edu.cn\}.}
 \thanks{K. Jia is with the Department of Electrical and Computer Engineering,
Faculty of Science and Technology, University of Macau, Macau 999078, China.
E-mail: \{kuijia@gmail.com\}.}
\thanks{D. Tao is with Centre for Quantum Computation \& Intelligent Systems,
 Faculty of Engineering \& Information Technology, University of Technology Sydney,
 235 Jones Street, Ultimo, NSW 2007, Australia.
 E-mail: \{dacheng.tao@uts.edu.au\}.}}

\markboth{}
{Shell \MakeLowercase{\textit{et al.}}: Bare Demo of IEEEtran.cls for Journals}
\maketitle

\begin{abstract}
Single image haze removal is a challenging ill-posed problem. Existing methods use various constraints/priors to get plausible dehazing solutions. The key to achieve haze removal is to estimate a medium transmission map for an input hazy image. In this paper, we propose a trainable end-to-end system called DehazeNet, for medium transmission estimation. DehazeNet takes a hazy image as input, and outputs its medium transmission map that is subsequently used to recover a haze-free image via atmospheric scattering model. DehazeNet adopts Convolutional Neural Networks (CNN) based deep architecture, whose layers are specially designed to embody the established assumptions/priors in image dehazing. Specifically, layers of Maxout units are used for feature extraction, which can generate almost all haze-relevant features. We also propose a novel nonlinear activation function in DehazeNet, called Bilateral Rectified Linear Unit (BReLU), which is able to improve the quality of recovered haze-free image.  We establish connections between components of the proposed DehazeNet and those used in existing methods. Experiments on benchmark images show that DehazeNet achieves superior performance over existing methods, yet keeps efficient and easy to use.
\end{abstract}


\IEEEpeerreviewmaketitle

\section{Introduction}\label{sec:Introduction}

\IEEEPARstart{H}{aze} is a traditional atmospheric phenomenon where dust, smoke and other dry particles obscure the clarity of the atmosphere. Haze causes issues in the area of terrestrial photography, where the light penetration of dense atmosphere may be necessary to image distant subjects. This results in the visual effect of a loss of contrast in the subject, due to the effect of light scattering through the haze particles. For these reasons, haze removal is desired in both consumer photography and computer vision applications.

Haze removal is a challenging problem because the haze transmission depends on the unknown depth which varies at different positions. Various techniques of image enhancement have been applied to the problem of removing haze from a single image, including histogram-based \cite{histogram}, contrast-based \cite{contrast} and saturation-based \cite{saturation}. In addition, methods using multiple images or depth information have also been proposed. For example, polarization based methods \cite{polarization} remove the haze effect through multiple images taken with different degrees of polarization. In \cite{multiimage1}, multi-constraint based methods are applied to multiple images capturing the same scene under different weather conditions. Depth-based methods \cite{depth2} require some depth information from user inputs or known 3D models. In practice, depth information or multiple hazy images are not always available.

Single image haze removal has made significant progresses recently, due to the use of better assumptions and priors. Specifically, under the assumption that the local contrast of the haze-free image is much higher than that of hazy image, a local contrast maximizing method \cite{maxcontrast} based on Markov Random Field (MRF) is proposed for haze removal. Although contrast maximizing approach is able to achieve impressive results, it tends to produce over-saturated images. In \cite{ica}, Independent Component Analysis (ICA) based on minimal input is proposed to remove the haze from color images, but the approach is time-consuming and cannot be used to deal with dense-haze images. Inspired by dark-object subtraction technique, Dark Channel Prior (DCP) \cite{dcp} is discovered based on empirical statistics of experiments on haze-free images, which shows at least one color channel has some pixels with very low intensities in most of non-haze patches. With dark channel prior, the thickness of haze is estimated and removed by the atmospheric scattering model. However, DCP loses dehazing quality in the sky images and is computationally intensive. Some improved algorithms are proposed to overcome these limitations. To improve dehazing quality, Kratz and Nishino \etal \cite{bayesian} model the image with a factorial MRF to estimate the scene radiance more accurately; Meng \etal \cite{bccr} propose an effective regularization dehazing method to restore the haze-free image by exploring the inherent boundary constraint. To improve computational efficiency, standard median filtering \cite{median}, median of median filter \cite{medianmedian}, guided joint bilateral filtering \cite{jointbilateral} and guided image filter \cite{guided} are used to replace the time-consuming soft matting \cite{softmatting}. In recent years, haze-relevant priors are investigated in machine learning framework. Tang \etal \cite{rf} combine four types of haze-relevant features with Random Forests to estimate the transmission. Zhu \etal \cite{cap} create a linear model for estimating the scene depth of the hazy image under color attenuation prior and learns the parameters of the model with a supervised method. Despite the remarkable progress, these state-of-the-art methods are limited by the very same haze-relevant priors or heuristic cues - they are often less effective for some images.

Haze removal from a single image is a difficult vision task. In contrast, the human brain can quickly identify the hazy area from the natural scenery without any additional information. One might be tempted to propose biologically inspired models for image dehazing, by following the success of bio-inspired CNNs for high-level vision tasks such as image classification \cite{imagenet}, face recognition \cite{facednn} and object detection \cite{objectdetection}. In fact, there have been a few (convolutional) neural network based deep learning methods that are recently proposed for low-level vision tasks of image restoration/reconstruction \cite{deconvolution,srcnn,dirtrain}. However, these methods cannot be directly applied to single image haze removal.

Note that apart from estimation of a global atmospheric light magnitude, the key to achieve haze removal is to recover an accurate medium transmission map. To this end, we propose DehazeNet, a trainable CNN based end-to-end system for medium transmission estimation. DehazeNet takes a hazy image as input, and outputs its medium transmission map that is subsequently used to recover the haze-free image by a simple pixel-wise operation. Design of DehazeNet borrows ideas from established assumptions/principles in image dehazing, while parameters of all its layers can be automatically learned from training hazy images. Experiments on benchmark images show that DehazeNet gives superior performance over existing methods, yet keeps efficient and easy to use. Our main contributions are summarized as follows.

\begin{enumerate}
\item DehazeNet is an end-to-end system. It directly learns and estimates the mapping relations between hazy image patches and their medium transmissions. This is achieved by special design of its deep architecture to embody established image dehazing principles.
\item We propose a novel nonlinear activation function in DehazeNet, called Bilateral Rectified Linear Unit\footnote{During the preparation of this manuscript (in December, 2015), we find that a nonlinear activation function called adjustable bounded rectifier is proposed in \cite{abr} (arXived in November, 2015), which is almost identical to BReLU. Adjustable bounded rectifier is motivated to achieve the objective of image recognition. In contrast, BReLU is proposed here to improve image restoration accuracy. It is interesting that we come to the same activation function from completely different initial objectives. This may also suggest the general usefulness of the proposed BReLU.} (BReLU). BReLU extends Rectified Linear Unit (ReLU) and demonstrates its significance in obtaining accurate image restoration. Technically, BReLU uses the bilateral restraint to reduce search space and improve convergence.
\item We establish connections between components of DehazeNet and those assumptions/priors used in existing dehazing methods, and explain that DehazeNet improves over these methods by automatically learning all these components from end to end.
\end{enumerate}

The remainder of this paper is organized as follows. In Section \ref{sec:related}, we review the atmospheric scattering model and haze-relevant features, which provides background knowledge to understand the design of DehazeNet. In Section \ref{sec:dehazecnn}, we present details of the proposed DehazeNet, and discuss how it relates to existing methods. Experiments are presented in Section \ref{sec:experiments}, before conclusion is drawn in Section \ref{sec:conclusion}.

\section{Related Works}\label{sec:related}
Many image dehazing methods have been proposed in the literature. In this section, we briefly review some important ones, paying attention to those proposing the atmospheric scattering model, which is the basic underlying model of image dehazing, and  those proposing useful assumptions for computing haze-relevant features.

\subsection{Atmospheric Scattering Model}\label{sec:asm}
\begin{figure*}
    \begin{minipage}{0.5\linewidth}
    \flushleft
    \subfigure[The process of imaging in hazy weather. The transmission attenuation $J\left(x\right)t\left(x\right)$ caused by the reduction in reflected energy, leads to low brightness intensity. The airlight $\alpha\left(1-t\left(x\right)\right)$ formed by the scattering of the environmental illumination, enhances the brightness and reduces the saturation.]{ \label{fig:subfig:round} 
    \includegraphics[width=0.95\linewidth]{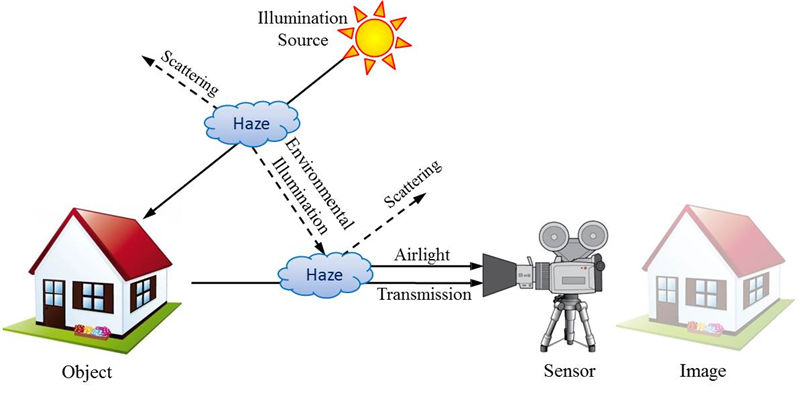}}
    \end{minipage}
    \begin{minipage}{0.5\linewidth}
    \flushright
    \subfigure[Atmospheric scattering model. The observed hazy image $I\left(x\right)$ is generated by the real scene $J\left(x\right)$, the medium transmission  $t\left(x\right)$ and the global atmospheric light $\alpha$.]{ \label{fig:subfig:maxout} 
    \includegraphics[width=0.95\linewidth]{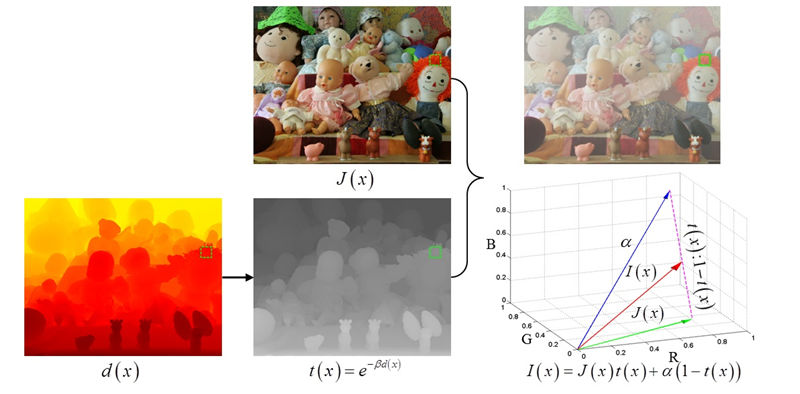}}
    \end{minipage}
    \caption{Imaging in hazy weather and atmospheric scattering model}
    \label{fig:atm}
\end{figure*}

To describe the formation of a hazy image, the atmospheric scattering model is first proposed by McCartney \cite{atmosphere}, which is further developed by Narasimhan and Nayar \cite{badweather,narasimhan}. The atmospheric scattering model can be formally written as
\begin{equation}\label{I}
I\left( x \right) = J\left( x \right)t\left( x \right) + \alpha\left( {1 - t\left( x \right)} \right),
\end{equation}
where $I\left( x \right)$ is the observed hazy image, $J\left( x \right)$ is the real scene to be recovered, $t\left( x \right)$ is the medium transmission, $\alpha$ is the global atmospheric light, and $x$ indexes pixels in the observed hazy image $I$. Fig. \ref{fig:atm} gives an illustration. There are three unknowns in equation \eqref{I}, and the real scene $J\left( x \right)$ can be recovered after $\alpha$ and $t\left( x \right)$ are estimated.

The medium transmission map $t\left(x\right)$ describes the light portion that is not scattered and reaches the camera. $t\left(x\right)$ is defined as
\begin{equation}\label{t}
t\left( x \right) = {e^{ - \beta d\left( x \right)}},
\end{equation}
where $d\left(x\right)$ is the distance from the scene point to the camera, and $\beta$ is the scattering coefficient of the atmosphere. Equation \eqref{t} suggests that when $d\left(x\right)$ goes to infinity, $t\left(x\right)$ approaches zero. Together with equation \eqref{I} we have
\begin{equation}\label{alpha}
\alpha = I\left( x \right),d\left( x \right) \to \inf
\end{equation}
In practical imaging of a distance view, $d\left( x \right)$ cannot be infinity, but rather be a long distance that gives a very low transmission $t_0$. Instead of relying on equation \eqref{alpha} to get the global atmospheric light $\alpha$, it is more stably estimated based on the following rule
\begin{equation}\label{Aa}
\alpha = \mathop {\max }\limits_{y \in \left\{ {x|t\left( x \right) \le {t_{0}}} \right\}} I\left( y \right)
\end{equation}

The discussion above suggests that \emph{to recover a clean scene (i.e., to achieve haze removal), it is the key to estimate an accurate medium transmission map.}

\subsection{Haze-relevant features}\label{sec:features}
Image dehazing is an inherently ill-posed problem. Based on empirical observations, existing methods propose various assumptions or prior knowledge that are utilized to compute intermediate haze-relevant features. Final haze removal can be achieved based on these haze-relevant features.
\subsubsection{Dark Channel}
The dark channel prior is based on the wide observation on outdoor haze-free images. In most of the haze-free patches, at least one color channel has some pixels whose intensity values are very low and even close to zero. The dark channel \cite{dcp} is defined as the minimum of all pixel colors in a local patch:
\begin{equation}\label{D}
D\left( x \right) = \mathop {\min }\limits_{y \in \Omega_r \left( x \right)} \left( {\mathop {\min }\limits_{c \in \left\{ {r,g,b} \right\}} {I^c}\left( y \right)} \right),
\end{equation}
where $I^c$ is a RGB color channel of $I$ and $\Omega_{r}\left(x\right)$ is a local patch centered at $x$ with the size of $r\times r$. The dark channel feature has a high correlation to the amount of haze in the image, and is used to estimate the medium transmission $t\left(x\right)\propto 1-D\left( x \right)$ directly.

\subsubsection{Maximum Contrast}
According to the atmospheric scattering, the contrast of the image is reduced by the haze transmission as $\sum\nolimits_x {\left\| {\nabla I\left( x \right)} \right\|}  = t\sum\nolimits_x {\left\| {\nabla J\left( x \right)} \right\|}  \le \sum\nolimits_x {\left\| {\nabla J\left( x \right)} \right\|}$ . Based on this observation, the local contrast \cite{maxcontrast} as the variance of pixel intensities in a $s\times s$ local patch $\Omega _s$ with respect to the center pixel, and the local maximum of local contrast values in a $r\times r$ region $\Omega _r$ is defined as:
\begin{equation}\label{C}
C\left( x \right) = \mathop {\max }\limits_{y \in {\Omega _r}\left( x \right)} \sqrt {\frac{1}{{\left| {{\Omega _s}\left( y \right)} \right|}}\sum\limits_{z \in {\Omega _s}\left( y \right)} {\left\| {I\left( z \right) - I\left( y \right)} \right\|}^2 },
\end{equation}
where $\left| {{\Omega _s}\left( y \right)} \right|$ is the cardinality of the local neighborhood. The correlation between the contrast feature and the medium transmission $t$ is visually obvious, so the visibility of the image is enhanced by maximizing the local contrast showed as \eqref{C}.

\subsubsection{Color Attenuation}
 The saturation $I^s\left(x\right)$ of the patch decreases sharply while the color of the scene fades under the influence of the haze, and the brightness value $I^v\left(x\right)$ increases at the same time producing a high value for the difference. According to the above color attenuation prior \cite{cap}, the difference between the brightness and the saturation is utilized to estimate the concentration of the haze:
\begin{equation}\label{A}
A\left( x \right) = {I^v}\left( x \right) - {I^s}\left( x \right),
\end{equation}
where $I^v\left(x\right)$ and $I^h\left(x\right)$ can be expressed in the HSV color space as ${I^v}\left( x \right) = \mathop {\max }\nolimits_{c \in \left\{ {r,b,g} \right\}} {I^c}\left( x \right)$ and ${I^s}\left( x \right) = {{\left( {\mathop {\max }\nolimits_{c \in \left\{ {r,b,g} \right\}} {I^c}\left( x \right) - \mathop {\min }\nolimits_{c \in \left\{ {r,b,g} \right\}} {I^c}\left( x \right)} \right)} \mathord{\left/
 {\vphantom {{\left( {\mathop {\max }\nolimits_{c \in \left\{ {r,b,g} \right\}} {I^c}\left( x \right) - \mathop {\min }\nolimits_{c \in \left\{ {r,b,g} \right\}} {I^c}\left( x \right)} \right)} {\mathop {\max }\nolimits_{c \in \left\{ {r,b,g} \right\}} {I^c}\left( x \right)}}} \right.
 \kern-\nulldelimiterspace} {\mathop {\max }\nolimits_{c \in \left\{ {r,b,g} \right\}} {I^c}\left( x \right)}}$. The color attenuation feature is proportional to the scene depth $d\left(x\right)\propto A\left(x\right)$, and is used for transmission estimation easily.

\subsubsection{Hue Disparity}
Hue disparity between the original image $I\left(x\right)$ and its semi-inverse image, $I_{si}\left( x \right) = {\max}\left[ {{I^c}\left( x \right),1 - {I^c}\left( x \right)} \right]$ with $c \in \left\{ {r,g,b} \right\}$, has been used to detect the haze. For haze-free images, pixel values in the three channels of their semi-inverse images will not all flip, resulting in large hue changes between $I_{si}\left(x\right)$ and $I\left(x\right)$. In \cite{semiinverse}, the hue disparity feature is defined:
\begin{equation}\label{H}
H\left( x \right) = \left| {I_{si}^h\left( x \right) - {I^h}\left( x \right)} \right|,
\end{equation}
where the superscript "$h$" denotes the hue channel of the image in HSV color space. According to \eqref{H}, the medium transmission $t\left(x\right)$ is in inverse propagation to $H\left(x\right)$.

\section{The Proposed DehazeNet}\label{sec:dehazecnn}

The atmospheric scattering model in Section \ref{sec:asm} suggests that estimation of the medium transmission map is the most important step to recover a haze-free image. To this end, we propose DehazeNet, a trainable end-to-end system that explicitly learns the mapping relations between raw hazy images and their associated medium transmission maps. In this section, we present layer designs of DehazeNet, and discuss how these designs are related to ideas in existing image dehazing methods.  The final pixel-wise operation to get a recovered haze-free image from the estimated medium transmission map will be presented in Section \ref{sec:experiments}.

\subsection{Layer Designs of DehazeNet}
\begin{figure*}[!t]
\centering
\includegraphics[width=0.95\linewidth]{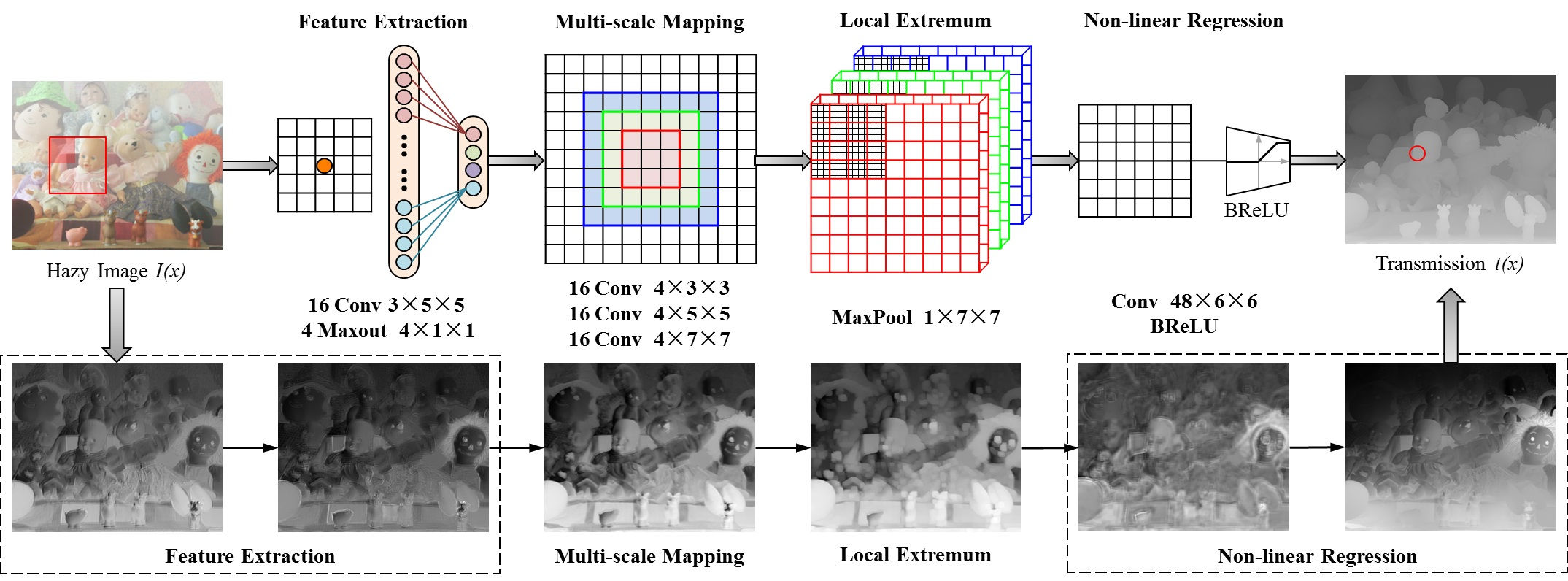}
\caption{The architecture of DehazeNet. DehazeNet conceptually consists of four sequential operations (feature extraction, multi-scale mapping, local extremum and non-linear regression), which is constructed by 3 convolution layers, a max-pooling, a Maxout unit and a BReLU activation function.}
\label{fig:hazenet}
\end{figure*}

The proposed DehazeNet consists of cascaded convolutional and pooling layers, with appropriate nonlinear activation functions employed after some of these layers. Fig. \ref{fig:hazenet} shows the architecture of DehazeNet. Layers and nonlinear activations of DehazeNet are designed to implement four sequential operations for medium transmission estimation, namely, feature extraction, multi-scale mapping, local extremum, and nonlinear regression. We detail these designs as follows.

\subsubsection{Feature Extraction}
To address the ill-posed nature of image dehazing problem, existing methods propose  various assumptions and based on these assumptions, they are able to extract haze-relevant features (e.g., dark channel, hue disparity, and color attenuation) densely over the image domain. Note that densely extracting these haze-relevant features is equivalent to convolving an input hazy image with appropriate filters, followed by nonlinear mappings. Inspired by extremum processing in color channels of those haze-relevant features, an unusual activation function called Maxout unit \cite{maxout} is selected as the non-linear mapping for dimension reduction. Maxout unit is a simple feed-forward nonlinear activation function used in multi-layer perceptron or CNNs. When used in CNNs, it generates a new feature map by taking a pixel-wise maximization operation over $k$ affine feature maps. Based on Maxout unit, we design the first layer of DehazeNet as follows
\begin{equation}\label{F1}
F_1^i\left(x\right) = \mathop {\max }\limits_{j \in \left[ {1,k} \right]} f_1^{i,j}\left(x\right), f_1^{i,j}={W_1^{i,j}{\rm{*}}I + B_1^{i,j}},
\end{equation}
where $\mathcal{W}_1=\{W_1^{i,j}\}_{(i,j)=(1,1)}^{(n_1,k)}$ and $\mathcal{B}_1=\{B_1^{i,j}\}_{(i,j)=(1,1)}^{(n_1,k)}$ represent the filters and biases respectively, and $\rm{*}$ denotes the convolution operation. Here, there are $n_1$ output feature maps in the first layer. $W_1^{i,j} \in { \mathbb{R}^{3 \times {f_1} \times {f_1}}}$ is one of the total $k\times n_1$ convolution filters, where 3 is the number of channels in the input image $I\left(x\right)$, and $f_1$ is the spatial size of a filter (detailed in Table \ref{tab:architectures}). Maxout unit maps each of $kn_1$-dimensional vectors into an $n_1$-dimensional one, and extracts the haze-relevant features by automatic learning rather than heuristic ways in existing methods.

\subsubsection{Multi-scale Mapping}
In \cite{rf}, multi-scale features have been proven effective for haze removal, which densely compute features of an input image at multiple spatial scales. Multi-scale feature extraction is also effective to achieve scale invariance. For example, the inception architecture in GoogLeNet \cite{googlenet} uses parallel convolutions with varying filter sizes, and better addresses the issue of aligning objects in input images, resulting in state-of-the-art performance in ILSVRC14 \cite{ilsvrc}. Motivated by these successes of multi-scale feature extraction,  we choose to use parallel convolutional operations in the second layer of DehazeNet, where size of any convolution filter is among {$3\times3$, $5\times5$ and $7\times7$}, and we use the same number of filters for these three scales. Formally, the output of the second layer is written as
\begin{equation}\label{F2}
F_2^i = W_2^{\left\lceil {i/3} \right\rceil ,\left( {i\backslash 3} \right)}{\rm{*}}{F_1} + {B_2}^{\left\lceil {i/3} \right\rceil ,\left( {i\backslash 3} \right)},
\end{equation}
where $\mathcal{W}_2=\{W_2^{p,q}\}_{(p,q)=(1,1)}^{(3,n_2/3)}$ and $\mathcal{B}_2=\{B_2^{p,q}\}_{(p,q)=(1,1)}^{(3,n_2/3)}$ contain $n_2$ pairs of parameters that is break up into 3 groups. $n_2$ is the output dimension of the second layer, and $i\in\left[1,n_2\right]$ indexes the output feature maps. $\left\lceil \right\rceil$ takes the integer upwardly and $\backslash$ denotes the remainder operation.

\subsubsection{Local Extremum}
To achieve spatial invariance, the cortical complex cells in the visual cortex receive responses from the simple cells for linear feature integration. Ilan \etal \cite{maxpool} proposed that spatial integration properties of complex cells can be described by a series of pooling operations. According to the classical architecture of CNNs \cite{lenet}, the neighborhood maximum is considered under each pixel to overcome local sensitivity. In addition, the local extremum is in accordance with the assumption that the medium transmission is locally constant, and it is commonly to overcome the noise of transmission estimation. Therefore, we use a local extremum operation in the third layer of DehazeNet.
\begin{equation}\label{F3}
{F_3^i}\left( x \right) = \mathop {\max }\limits_{y \in {\Omega\left(x\right)}} {F_2^i}\left(y\right),
\end{equation}
where $\Omega\left(x\right)$ is an $f_3\times f_3$ neighborhood centered at $x$, and the output dimension of the third layer $n_3=n_2$. In contrast to max-pooling in CNNs, which usually reduce resolutions of feature maps, the local extremum operation here is densely applied to every feature map pixel, and is able to preserve resolution for use of image restoration.

\subsubsection{Non-linear Regression}
\begin{figure}[!t]
\centering
\subfigure[ReLU]{ \label{fig:drelu:relu} 
    \includegraphics[width=0.4\linewidth]{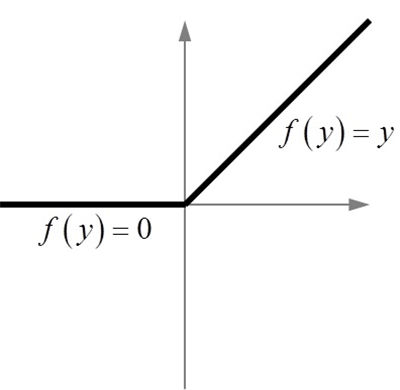}}
    \subfigure[BReLU]{ \label{fig:drelu:drelu} 
    \includegraphics[width=0.42\linewidth]{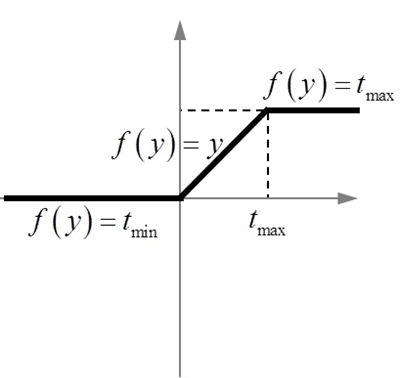}}
\caption{Rectified Linear Unit (ReLU) and Bilateral Rectified Linear Unit (BReLU)}
\label{fig:drelu}
\end{figure}

Standard choices of nonlinear activation functions in deep networks include Sigmoid \cite{function} and Rectified Linear Unit (ReLU). The former one is easier to suffer from vanishing gradient, which may lead to slow convergence or poor local optima in networks training. To overcome the problem of vanishing gradient, ReLU is proposed \cite{relu} which offers sparse representations. However, ReLU is designed for classification problems and not perfectly suitable for the regression problems such as image restoration. In particular, ReLU inhibits values only when they are less than zero. It might lead to response overflow especially in the last layer, because for image restoration, the output values of the last layer are supposed to be both lower and upper bounded in a small range.  To this end, we propose a Bilateral Rectified Linear Unit (BReLU) activation function, shown in Fig. \ref{fig:drelu}, to overcome this limitation. Inspired by Sigmoid and ReLU, BReLU as a novel linear unit keeps bilateral restraint and local linearity. Based on the proposed BReLU, the feature map of the fourth layer is defined as
\begin{equation}\label{F4}
{F_4} = \min \left( {t_{\max},\max \left( {t_{\min},{W_4}*{F_3} + {B_4}} \right)} \right)
\end{equation}
Here $\mathcal{W}_4=\{W_4\}$ contains a filter with the size of $n_3\times f_4\times f_4$, $\mathcal{B}_4=\{B_4\} $ contains a bias, and $t_{\min,\max}$ is the marginal value of BReLU ($t_{\min}=0$ and $t_{\max}=1$ in this paper). According to \eqref{F4}, the gradient of this activation function can be shown as
\begin{equation}\label{dF4}
\frac{{\partial {F_4}\left( x \right)}}{{\partial F_3}} = \left\{ \begin{array}{l}
\dfrac{{\partial {F_4}\left( x \right)}}{{\partial F_3}},t_{\min} \le {F_4}\left( x \right) < t_{\max}\\
0,otherwise
\end{array} \right.
\end{equation}

The above four layers are cascaded together to form a CNN based trainable end-to-end system, where filters and biases associated with convolutional layers are network parameters to be learned. We note that designs of these layers can be connected with expertise in existing image dehazing methods, which we specify in the subsequent section.

\subsection{Connections with Traditional Dehazing Methods}\label{sec:relationship}

\begin{figure}
    \centering
    \subfigure[Opposite filter]{ \label{fig:subfig:opposite} 
    \includegraphics[width=0.22\linewidth]{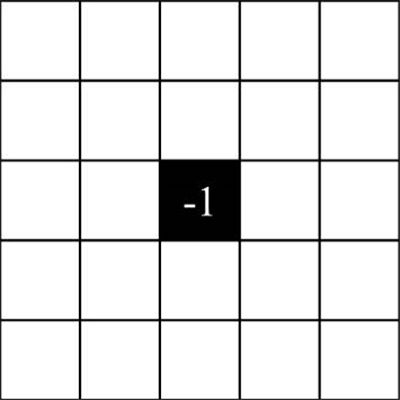}}
    \subfigure[All-pass filter]{ \label{fig:subfig:allpass} 
    \includegraphics[width=0.22\linewidth]{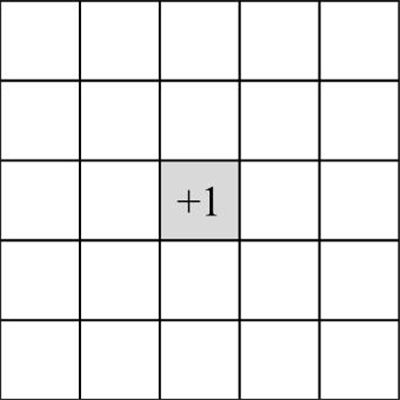}}
    \subfigure[Round filter]{ \label{fig:subfig:round} 
    \includegraphics[width=0.22\linewidth]{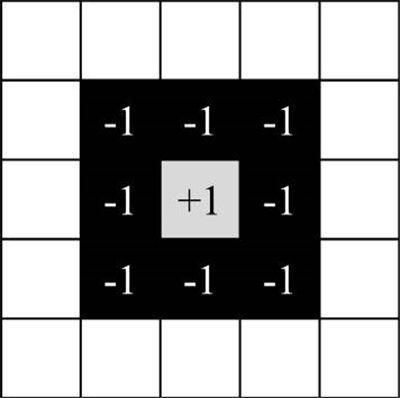}}
    \subfigure[Maxout]{ \label{fig:subfig:maxout} 
    \includegraphics[width=0.22\linewidth]{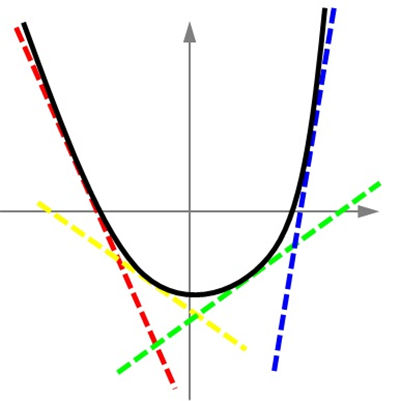}}

    \subfigure[The actual kernels learned from DehazeNet]{
    \label{fig:subfig:kernel}
    \begin{minipage}[b]{1.0\linewidth}
    \centering
    \includegraphics[width=0.1\linewidth]{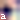}
    \includegraphics[width=0.1\linewidth]{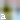}
    \includegraphics[width=0.1\linewidth]{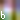}
    \includegraphics[width=0.1\linewidth]{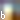}
    \includegraphics[width=0.1\linewidth]{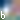}
    \includegraphics[width=0.1\linewidth]{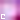}
    \includegraphics[width=0.1\linewidth]{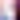}
    \includegraphics[width=0.1\linewidth]{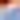}
    \\[0.1cm]
    \includegraphics[width=0.1\linewidth]{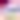}
    \includegraphics[width=0.1\linewidth]{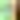}
    \includegraphics[width=0.1\linewidth]{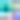}
    \includegraphics[width=0.1\linewidth]{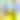}
    \includegraphics[width=0.1\linewidth]{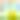}
    \includegraphics[width=0.1\linewidth]{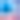}
    \includegraphics[width=0.1\linewidth]{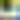}
    \includegraphics[width=0.1\linewidth]{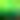}
    \end{minipage}
    }
    \caption{Filter weight and Maxout unit in the first layer operation $F_1$}
    \label{fig:W1}
\end{figure}

The first layer feature $F_1$ in DehazeNet is designed for haze-relevant feature extraction. Take dark channel feature \cite{dcp} as an example. If the weight $W_1$ is an opposite filter (sparse matrices with the value of -1 at the center of one channel, as in Fig. \ref{fig:subfig:opposite}) and $B_1$ is a unit bias, then the maximum output of the feature map is equivalent to the minimum of color channels, which is similar to dark channel \cite{dcp} (see Equation \eqref{D}). In the same way, when the weight is a round filter as Fig. \ref{fig:subfig:round}, $F_1$ is similar to the maximum contrast \cite{maxcontrast} (see Equation \eqref{C}); when $\mathcal{W}_1$ includes all-pass filters and opposite filters, $F_1$ is similar to the maximum and minimum feature maps, which are atomic operations of the color space transformation from RGB to HSV, then the color attenuation \cite{cap} (see Equation \eqref{A}) and hue disparity \cite{semiinverse} (see Equation \eqref{H}) features are extracted. In conclusion, upon success of filter learning shown in Fig. \ref{fig:subfig:kernel}, almost all haze-relevant features can be potentially extracted from the first layer of DehazeNet. On the other hand, Maxout activation functions can be considered as piece-wise linear approximations to arbitrary convex functions. In this paper, we choose the maximum across four feature maps ($k=4$) to approximate an arbitrary convex function, as shown in Fig. \ref{fig:subfig:maxout}.

  White-colored objects in an image are similar to heavy haze scenes that are usually with high values of brightness and low values of saturation. Therefore, almost all the haze estimation models tend to consider the white-colored scene objects as being distant, resulting in inaccurate estimation of the medium transmission. Based on the assumption that the scene depth is locally constant, local extremum filter is commonly to overcome this problem \cite{dcp,cap,maxcontrast}. In DehazeNet, local maximum filters of the third layer operation remove the local estimation error. Thus the direct attenuation term $J\left(x\right)t\left(x\right)$ can be very close to zero when the transmission $t\left(x\right)$ is close to zero. The directly recovered scene radiance $J\left(x\right)$ is prone to noise. In DehazeNet, we propose BReLU to restrict the values of transmission between $t_{\min}$ and $t_{\max}$, thus alleviating the noise problem. Note that BReLU is equivalent to the boundary constraints used in traditional methods \cite{dcp,cap}.

\subsection{Training of DehazeNet}

\subsubsection{Training Data}
It is in general costly to collect a vast amount of labelled data for training deep models \cite{imagenet}. For training of DehazeNet, it is even more difficult as the pairs of hazy and haze-free images of natural scenes (or the pairs of hazy images and their associated medium transmission maps) are not massively available. Instead, we resort to synthesized training data based on the physical haze formation model \cite{rf}.

More specifically, we synthesize training pairs of hazy and haze-free image patches based on two assumptions \cite{rf}: first, image content is independent of medium transmission (the same image content can appear at any depths of scenes); second, medium transmission is locally constant (image pixels in a small patch tend to have similar depths). These assumptions suggest that we can assume an arbitrary transmission for an individual image patch. Given a haze-free patch $J^P\left(x\right)$, the atmospheric light $\alpha$, and a random transmission $t\in\left(0,1\right)$, a hazy patch is synthesized as $I^P\left(x\right) = J^P\left(x\right)t + \alpha\left(1-t\right)$. To reduce the uncertainty in variable learning, atmospheric light $\alpha$ is set to 1.

\begin{figure}[!t]
\centering
\subfigure{
\includegraphics[width=0.19\linewidth]{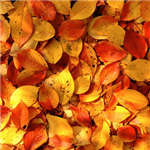}
\includegraphics[width=0.19\linewidth]{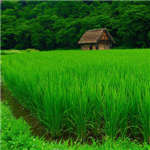}
\includegraphics[width=0.19\linewidth]{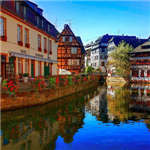}
\includegraphics[width=0.19\linewidth]{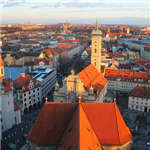}
\includegraphics[width=0.19\linewidth]{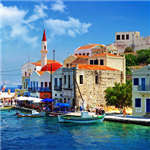}}
\subfigure{
\includegraphics[width=0.19\linewidth]{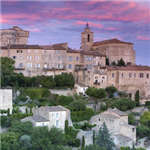}
\includegraphics[width=0.19\linewidth]{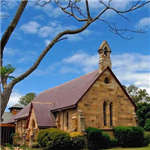}
\includegraphics[width=0.19\linewidth]{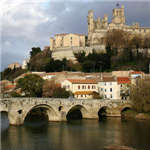}
\includegraphics[width=0.19\linewidth]{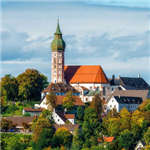}
\includegraphics[width=0.19\linewidth]{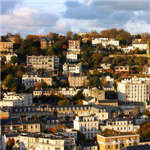}}
\caption{Example haze-free training images collected from the Internet}
\label{fig:hazefree}
\end{figure}

 In this work, we collect haze-free images from the Internet, and randomly sample from them patches of size $16\times16$. Different from \cite{rf}, these haze-free images include not only those capturing people's daily life, but also those of natural and city landscapes, since we believe that this variety of training samples can be learned into the filters of DehazeNet. Fig. \ref{fig:hazefree} shows examples of our collected haze-free images.

\subsubsection{Training Method}\label{sec:train}
In the DehazeNet, supervised learning requires the mapping relationship $\mathcal{F}$ between RGB value and medium transmission. Network parameters $\Theta = \left\{\mathcal{W}_1,\mathcal{W}_2,\mathcal{W}_4,\mathcal{B}_1,\mathcal{B}_2,\mathcal{B}_4\right\}$ are achieved through minimizing the loss function between the training patch $I^P\left(x\right)$ and the corresponding ground truth medium transmission $t$. Given a set of hazy image patches and their corresponding medium transmissions, where hazy patches are synthesized from haze-free patches as described above, we use Mean Squared Error (MSE) as the loss function:
\begin{equation}\label{loss}
L\left( \Theta  \right) = \frac{1}{N}\sum\limits_{i = 1}^N {{{\left\| {\mathcal{F}\left( {I_i^P;\Theta } \right) - {t_i}} \right\|}^2}}
\end{equation}

Stochastic gradient descent (SGD) is used to train DehazeNet. We implement our model using the \emph{Caffe} package \cite{caffe}. Detailed configurations and parameter settings of our proposed DehazeNet (as shown in Fig.  \ref{fig:hazenet}) are summarized in Table  \ref{tab:architectures}, which includes 3 convolutional layers and 1 max-pooling layer, with Maxout and BReLU activations used respectively after the first and last convolutional operations.
\begin{table}
\center
\caption{The architectures of the DehazeNet model}
\begin{tabular}{|c|c|c|c|c|c|}
  \hline
  \multirow{2}{*}{Formulation} & \multirow{2}{*}{Type} & \multirow{2}{*}{Input Size} & Num & Filter& \multirow{2}{*}{Pad}\\
  & & & $n$ & $f\times f$& \\
  \hline
  \hline
  \multirow{2}{*}{\minitab[c]{Feature\\Extraction}} & Conv & $3\times16\times16$ & 16 & $5\times5$ & 0 \\
  & Maxout & $16\times12\times12$ & 4 & -- & 0\\
  \hline
  \hline
  \multirow{3}{*}{\minitab[c]{Multi-scale\\Mapping}} & \multirow{3}{*}{Conv} & \multirow{3}{*}{$4\times12\times12$} & 16 & $3\times3$ & 1 \\
  & & & 16 & $5\times5$ & 2 \\
  & & & 16 & $7\times7$ & 3 \\
  \hline\hline
  Local Extremum & Maxpool & $48\times12\times12$ & -- & $7\times7$ & 0\\
  \hline\hline
  \multirow{2}{*}{\minitab[c]{Non-linear\\Regression}} & Conv & $48\times6\times6$ & 1 & $6\times6$ & 0 \\
  & BReLU & $1\times1$ & 1 & -- & 0\\
  \hline
\end{tabular}
\label{tab:architectures}
\end{table}

\section{Experiments}\label{sec:experiments}
To verify the architecture of DehazeNet, we analyze its convergence and compare it with the state-of-art methods, including FVR \cite{fvr}, DCP \cite{dcp}, BCCR \cite{bccr}, ATM \cite{atm}, RF \cite{rf}, BPNN \cite{bpnn}, RF \cite{rf} and CAP \cite{cap}.

Regarding the training data, 10,000 haze-free patches are sampled randomly from the images collected from the Internet. For each patch, we uniformly sample 10 random transmissions $t\in\left(0,1\right)$ to generate 10 hazy patches. Therefore, a total of 100,000 synthetic patches are generated for DehazeNet training. In DehazeNet, the filter weights of each layer are initialized by drawing randomly from a Gaussian distribution (with mean value $\mu=0$ and standard deviation $\sigma=0.001$), and the biases are set to 0. The learning rate decreases by half from 0.005 to 3.125e-4 every 100,000 iterations. Based on the parameters above, DehazeNet is trained (in 500,000 iterations with a batch-size of 128) on a PC with Nvidia GeForce GTX 780 GPU.

Based on the transmission estimated by DehazeNet and the atmospheric scattering model, haze-free images are restored as traditional methods. Because of the local extremum in the third layer, the blocking artifacts appear in the transmission map obtained from DehazeNet. To refine the transmission map, guided image filtering \cite{guided} is used to smooth the image. Referring to Equation \eqref{Aa}, the boundary value of 0.1 percent intensity is chosen as $t_0$ in the transmission map, and we select the highest intensity pixel in the corresponding hazy image $I\left(x\right)$ among $x\in\{y|t\left(y\right)\leq t_0\}$ as the atmospheric light $\alpha$. Given the medium transmission $t\left(x\right)$ and the atmospheric light $\alpha$, the haze-free image $J\left(x\right)$ is recovered easily. For convenience, Equation \eqref{I} is rewritten as follows:
\begin{equation}\label{J}
J\left( x \right) = \frac{{I\left( x \right) - \alpha \left( {1 - t\left( x \right)} \right)}}{{t\left( x \right)}}
\end{equation}

Although DehazeNet is based on CNNs, the lightened network can effectively guarantee the realtime performance, and runs without GPUs. The entire dehazing framework is tested in MATLAB 2014A only with a CPU (Intel i7 3770, 3.4GHz), and it processes a $640\times480$ image with approximately 1.5 seconds.

\subsection{Model and performance}
In DehazeNet, there are two important layers with special design for transmission estimation, feature extraction $F_1$ and non-linear regression $F_4$. To proof the effectiveness of DehazeNet, two traditional CNNs (SRCNN \cite{srcnneccv} and CNN-L \cite{srcnn}) with the same number of 3 layers are regarded as baseline models. The number of parameters of DehazeNet, SRCNN, and CNN-L is 8,240, 18,400, and 67,552 respectively.

\subsubsection{Maxout unit in feature extraction $F_1$}
The activation unit in $F_1$ is a non-linear dimension reduction to approximate traditional haze-relevant features extraction. In the field of image processing, low-dimensional mapping is a core procedure for discovering principal attributes and for reducing pattern noise. For example, PCA \cite{pcanet} and LDA \cite{lda} as classical linear low-dimensional mappings are widely used in computer vision and data mining. In \cite{srcnn}, a non-linear sparse low-dimensional mapping with ReLU is used for high-resolution reconstruction. As an unusual low-dimensional mapping, Maxout unit maximizes feature maps to discover the prior knowledge of the hazy images. Therefore, the following experiment is designed to confirm the validity of Maxout unit. According to \cite{srcnn}, linear unit maps a 16-dimensional vector into a 4-dimensional vector, which is equivalent to applying 4 filters with a size of $16\times1\times1$. In addition, the sparse low-dimensional mapping is connecting ReLU to the linear unit.

\begin{figure}[!t]
\centering
\includegraphics[width=1.0\linewidth]{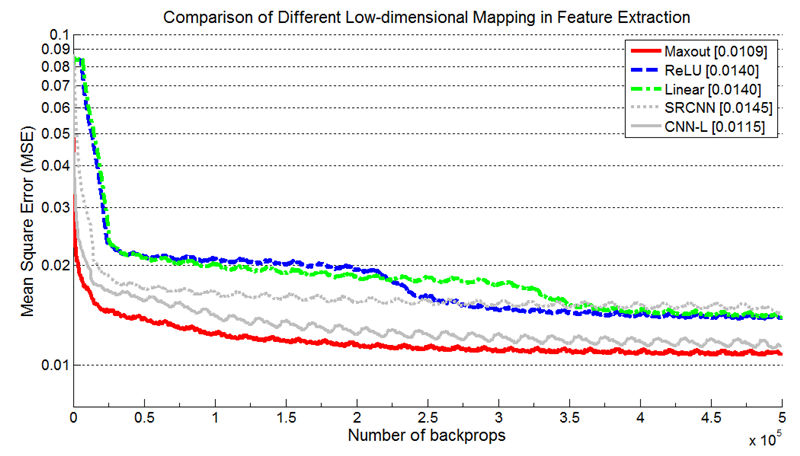}
\caption{The training process with different low-dimensional mapping in $F_1$}
\label{fig:f1}
\end{figure}

Fig. \ref{fig:f1} presents the training process of DehazeNet with Maxout unit, compared with ReLU and linear unit. We observe in Fig. \ref{fig:f1} that the speed of convergence for Maxout network is faster than that for ReLU and linear unit. In addition, the values in the bracket present the convergent result, and the performance of Maxout is improved by approximately 0.30e-2 compared with ReLU and linear unit. The reason is that Maxout unit provides the equivalent function of almost all of haze-relevant features, and alleviates the disadvantage of the simple piecewise functions such as ReLU.

\subsubsection{BReLU in non-linear regression $F_4$}
BReLU is a novel activation function that is useful for image restoration and reconstruction. Inspired by ReLU and Sigmoid, BReLU is designed with bilateral restraint and local linearity. The bilateral restraint applies a priori constraint to reduce the solution space scale; the local linearity overcomes the gradient vanishing to gain better precision. In the contrast experiment, ReLU and Sigmoid are used to take the place of BReLU in the non-linear regression layer. For ReLU, $F_4$ can be rewritten as ${F_4} = \max \left( {0,{W_4} * {F_3} + {B_4}} \right)$, and for Sigmoid, it can be rewritten as ${F_4} = {1 \mathord{\left/
 {\vphantom {1 {\left( {1 + \exp \left( { - {W_4} * {F_3} - {B_4}} \right)} \right)}}} \right.
 \kern-\nulldelimiterspace} {\left( {1 + \exp \left( { - {W_4} * {F_3} - {B_4}} \right)} \right)}}$.

\begin{figure}[!t]
\centering
\includegraphics[width=1.0\linewidth]{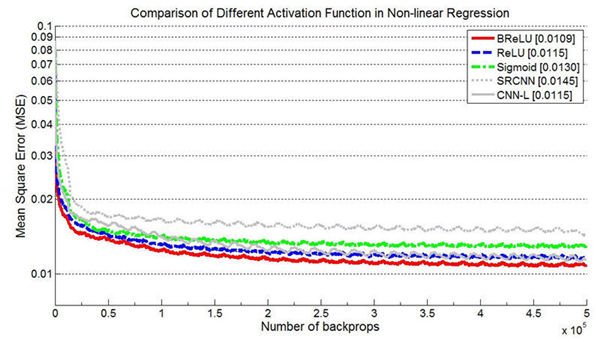}
\caption{The training process with different activation function in $F_4$}
\label{fig:f4}
\end{figure}

\begin{figure*}[!t]
\centering
\includegraphics[width=0.3\linewidth]{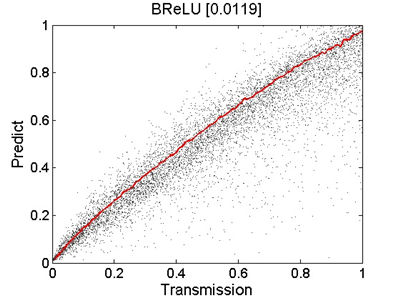}
\includegraphics[width=0.3\linewidth]{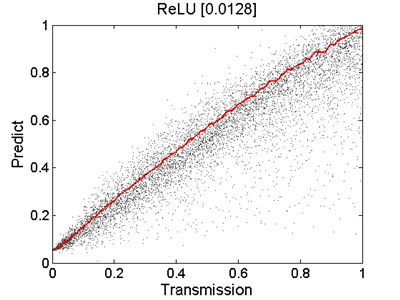}
\includegraphics[width=0.3\linewidth]{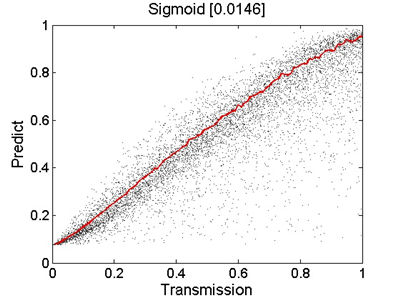}
\caption{The plots between predicted and truth transmission on different activation function in the non-linear regression $F_4$}
\label{fig:f4plot}
\end{figure*}

Fig. \ref{fig:f4} shows the training process using different activation functions in $F_4$. BReLU has a better convergence rate than ReLU and Sigmoid, especially during the first 50,000 iterations. The convergent precisions show that the performance of BReLU is improved approximately 0.05e-2 compared with ReLU and by 0.20e-2 compared with Sigmoid. Fig. \ref{fig:f4plot} plots the predicted transmission versus the ground truth transmission on the test patches. Clearly, the predicted transmission centers around the 45 degree line in BReLU result. However, the predicted transmission of ReLU is always higher than the true transmission, and there are some predicted transmissions over the limit value $t_{\max}=1$. Due to the curvature of Sigmoid function, the predicted transmission is far away from the true transmission, closing to 0 and 1. The MSE on the test set of BReLU is 1.19e-2, and that of ReLU and Sigmoid are 1.28e-2 and 1.46e-2, respectively.

\subsection{Filter number and size}

To investigate the best trade-off between performance and parameter size, we progressively modify the parameters of DehazeNet. Based on the default settings of DehazeNet, two experiments are conducted: (1) one is with a larger filter number, and (2) the other is with a different filter size. Similar to Sec. \ref{sec:train}, these models are trained on the same dataset and Table \ref{tab:filter} shows the training/testing MSEs with the corresponding parameter settings.
\begin{table}
\center
\caption{The results of using different filter number or size in DehazeNet ($\times10^{-2}$)}
\begin{tabular}{|c|c|c|c|c|}
  \hline
  Filter & Architecture & Train MSE & Test MSE & \#Param\\
  \hline
  \hline
  \multirow{3}{*}{\tabincell{c}{Number\\($n_1$-$n_2$)}} &\textbf{4-(16$\times$3)} & \textbf{1.090} & \textbf{1.190} &\hspace{0.5em}\textbf{8,240}\\
  &8-($32\times3$) & 0.972 & 1.138 &27,104\\
  &16-($64\times3$) & 0.902 & 1.112 &96,704\\
  \hline
  \hline
  \multirow{4}{*}{\tabincell{c}{$F_2$ Size\\($f_1$-$f_2$-$f_3$-$f_4$)}} &5-3-7-6& 1.184 & 1.219 & \hspace{0.5em}4,656\\
  &5-5-7-6& 1.133 & 1.225 &\hspace{0.5em}7,728\\
  &5-7-7-6& 1.021 & 1.184 & 12,336\\
  &\textbf{5-M-7-6}& \textbf{1.090} & \textbf{1.190} &\hspace{0.5em}\textbf{8,240}\\
  \hline
  \hline
  \multirow{3}{*}{\tabincell{c}{$F_4$ Size\\($f_1$-$f_2$-$f_3$-$f_4$)}} &5-M-6-7 & 1.077 & 1.192 & \hspace{0.5em}8,864\\
  &\textbf{5-M-7-6} & \textbf{1.090} & \textbf{1.190} &\hspace{0.5em}\textbf{8,240}\\
  &5-M-8-5 & 1.103 & 1.201 &\hspace{0.5em}7,712\\
  \hline
\end{tabular}
\label{tab:filter}
\end{table}

In general, the performance would improve when increasing the network width. It is clear that superior performance could be achieved by increasing the number of filter. However, if a fast dehazing speed is desired, a small network is preferred, which could still achieve better performance than other popular methods. In this paper, the lightened network is adopted in the following experiments.

In addition, we examine the network sensitivity to different filter sizes. The default network setting, whose specifics are shown in Table \ref{tab:architectures}, is denoted as 5-M-7-6. We first analyze the effects of varying filter sizes in the second layer $F_2$. Table \ref{tab:filter} indicates that a reasonably larger filter size in $F_2$ could grasp richer structural information, which in turn leads to better results. The multi-scale feature mapping with the filter sizes of 3/5/7 is also adopted in F2 of DehazeNet, which achieves similar testing MSE to that of the single-scale case of $7\times7$ filter. Moreover, we demonstrate in Section \ref{sec:evaluation} that the multi-scale mapping is able to improve the scale robustness.

We further examine networks with different filter sizes in the third and fourth layer. Keeping the same receptive field of network, the filter sizes in $F_3$ and $F_4$ are adjusted  simultaneously. It is showed that the larger filter size of non-linear regression $F_4$ enhances the fitting effect, but may lead to over-fitting. The local extremum in $F_3$ could improve the robustness on testing dataset. Therefore, we find the best filter setting of $F_3$ and $F_4$ as 5-M-7-6 in DehazeNet.

\subsection{Quantitative results on synthetic patches}

 In recent years, there are three methods based on learning framework for haze removal. In \cite{cap}, dehazing parameters are learned by a linear model, to estimate the scene depth under Color Attenuation Prior (CAP). A back propagation neural network (BPNN) \cite{bpnn} is used to mine the internal link between color and depth from the training samples. In \cite{rf}, Random Forests (RF) are used to investigate haze-relevant features for haze-free image restoration. All of the above methods and DehazeNet are trained with the same method as RF. According to the testing measure of RF, 2000 image patches are randomly sampled from haze-free images with 10 random transmission $t\in \left(0,1\right)$ to generate 20,000 hazy patches for testing. We run DehazeNet and CAP on the same testing dataset to measure the mean squared error (MSE) between the predicted transmission and true transmission. DCP \cite{dcp} is a classical dehazing method, which is used as a comparison baselines.
\begin{table}
\center
\caption{MSE between predicted transmission and ground truth on synthetic patches}
\begin{tabular}{c|c|c|c|c|c}
  \hline  \hline
  Methods&DCP \cite{dcp}&BPNN \cite{bpnn}&CAP \cite{cap}&RF \cite{rf}&DehazeNet\\
  \hline
  MSE($\times10^{-2}$)&3.18&4.37&3.32&1.26&\textbf{1.19}\\
  \hline \hline
\end{tabular}
\label{tab:syntheticpatches}
\end{table}

Table \ref{tab:syntheticpatches} shows the MSE between predicted transmissions and truth transmissions on the testing patches. DehazeNet achieves the best state-of-the-art score, which is 1.19e-2; the MSE between our method and the next state-of-art result (RF \cite{rf}) in the literature is 0.07e-2. Because in \cite{rf}, the feature values of patches are sorted to break the correlation between the haze-relevant features and the image content. However, the content information concerned by DehazeNet is useful for the transmission estimation of the sky region and white objects. Moreover, CAP \cite{cap} achieves satisfactory results in follow-on experiments but poor performance in this experiment, due to some outlier values (greater than 1 or less than 0) in the linear regression model.

\subsection{Quantitative results on synthetic images}\label{sec:evaluation}

\begin{figure}[!t]
\centering
\subfigure{
\begin{minipage}[b]{1.0\linewidth}
\includegraphics[width=0.23\linewidth]{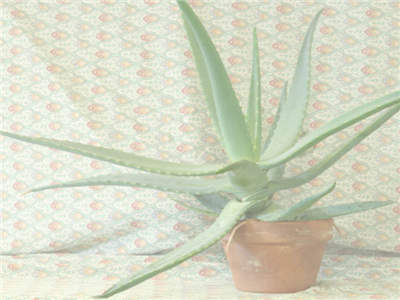}
\includegraphics[width=0.23\linewidth]{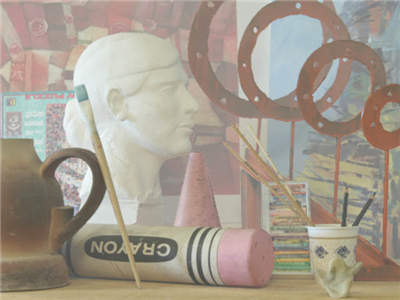}
\includegraphics[width=0.23\linewidth]{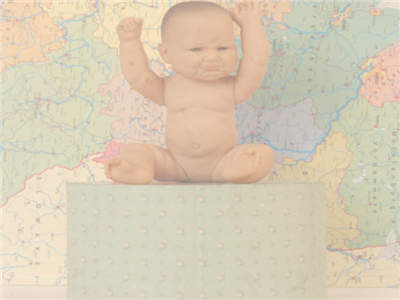}
\includegraphics[width=0.23\linewidth]{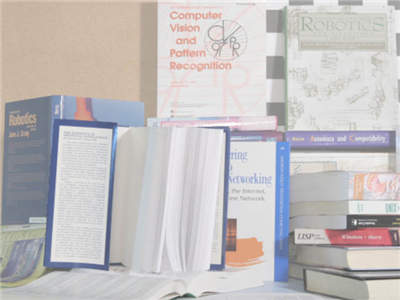}\\
\includegraphics[width=0.23\linewidth]{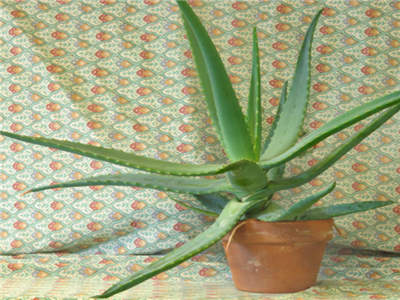}
\includegraphics[width=0.23\linewidth]{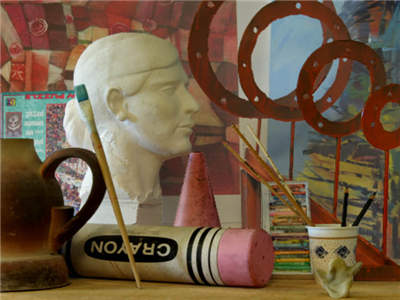}
\includegraphics[width=0.23\linewidth]{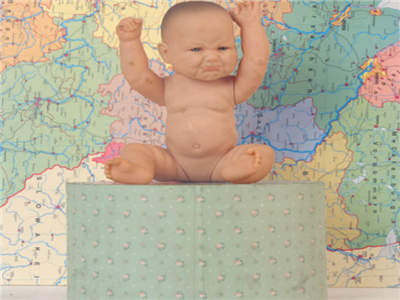}
\includegraphics[width=0.23\linewidth]{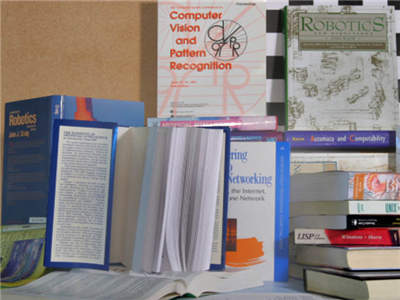}
\end{minipage}}
\subfigure{
\begin{minipage}[b]{1.0\linewidth}
\includegraphics[width=0.23\linewidth]{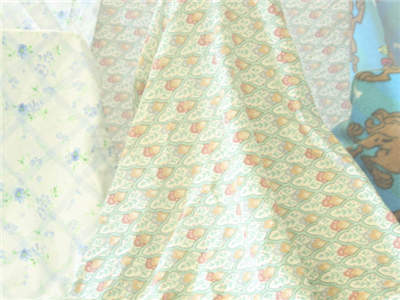}
\includegraphics[width=0.23\linewidth]{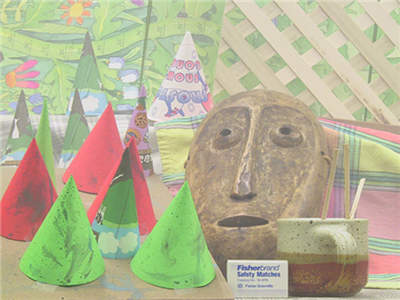}
\includegraphics[width=0.23\linewidth]{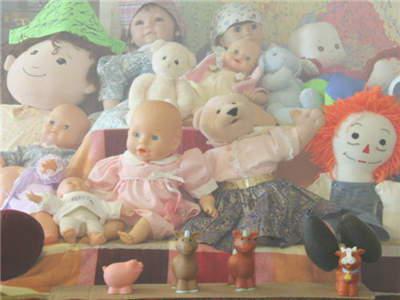}
\includegraphics[width=0.23\linewidth]{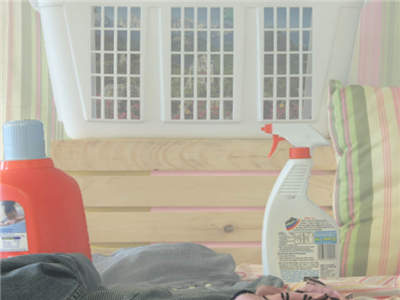}\\
\includegraphics[width=0.23\linewidth]{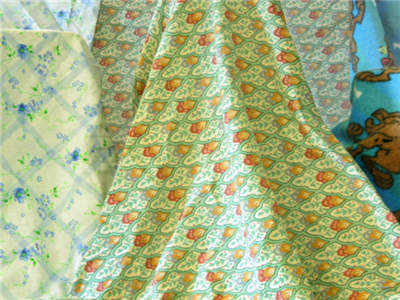}
\includegraphics[width=0.23\linewidth]{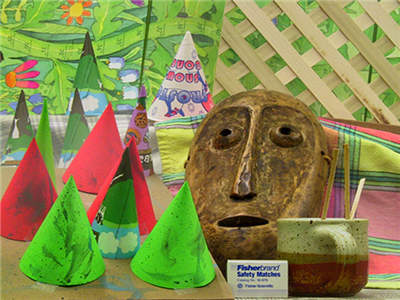}
\includegraphics[width=0.23\linewidth]{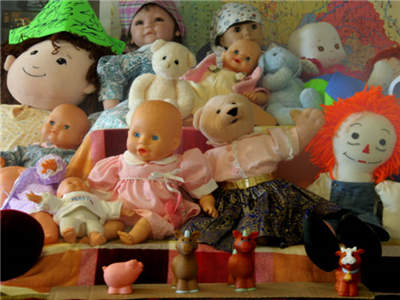}
\includegraphics[width=0.23\linewidth]{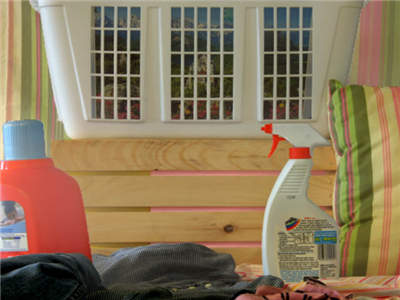}
\end{minipage}
}
\subfigure{
\begin{minipage}[b]{1.0\linewidth}
\includegraphics[width=0.23\linewidth]{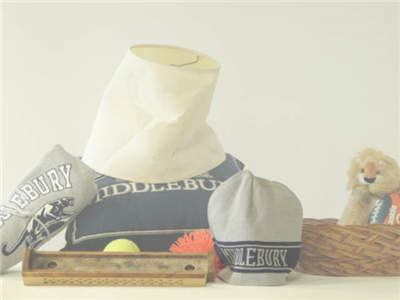}
\includegraphics[width=0.23\linewidth]{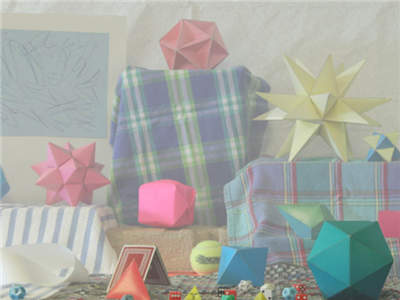}
\includegraphics[width=0.23\linewidth]{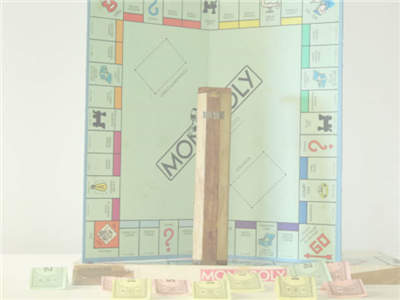}
\includegraphics[width=0.23\linewidth]{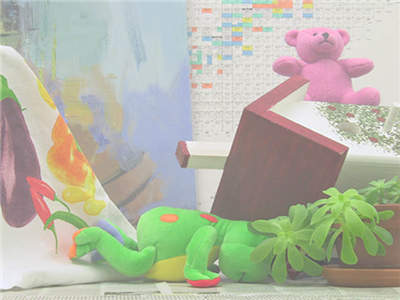}\\
\includegraphics[width=0.23\linewidth]{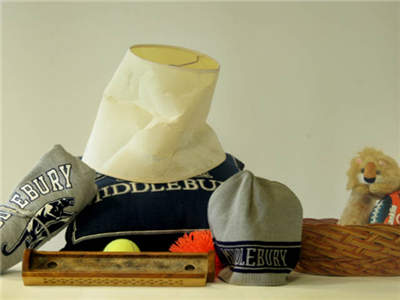}
\includegraphics[width=0.23\linewidth]{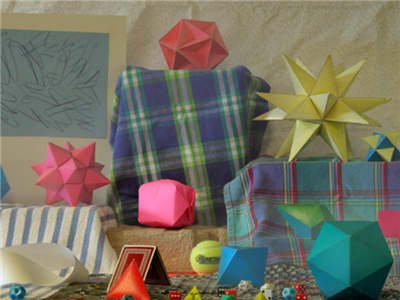}
\includegraphics[width=0.23\linewidth]{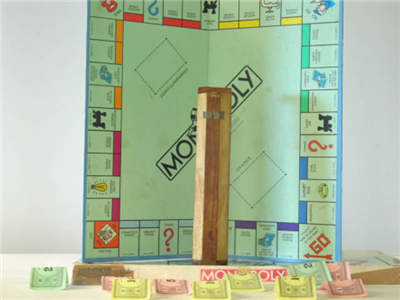}
\includegraphics[width=0.23\linewidth]{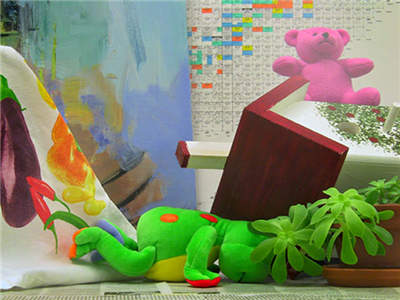}
\end{minipage}
}
\caption{Synthetic images based on Middlebury Stereo Datasets and DehazeNet results}
\label{fig:dataset}
\end{figure}

To verify the effectiveness on complete images, DehazeNet is tested on synthesized hazy images from stereo images with a known depth map $d\left(x\right)$, and it is compared with DCP \cite{dcp}, FVR \cite{fvr}, BCCR \cite{bccr}, ATM \cite{atm}, CAP \footnote{The results outside the parenthesis are run with the code implemented by authors \cite{cap}, and the results in the parenthesis are re-implemented by us.\label{fn:cap}}\cite{cap}, and RF \cite{rf}. There are 12 pairs of stereo images collected in Middlebury Stereo Datasets (2001-2006) \cite{middlebury01,middlebury02,middlebury03}. In Fig. \ref{fig:dataset}, the hazy images are synthesized from the haze-free stereo images based on \eqref{I}, and they are restored to haze-free images by DehazeNet.

To quantitatively assess these methods, we use a series of evaluation criteria in terms of the difference between each pair of haze-free image and dehazing result. Apart from the widely used mean square error (MSE) and the structural similarity (SSIM) \cite{ssim} indices, we used additional evaluation matrices, namely peak signal-to-noise ratio (PSNR) and weighted peak signal-to-noise ratio (WPSNR) \cite{wsnr}. We define one-pass evaluation (OPE) as the conventional method, which we run with standard parameters and report the average measure for the performance assessment. In Table \ref{tab:synthetic_images}, DehazeNet is compared with six state-of-the-art methods on all of the hazy images by OPE (hazy images are synthesized with the single scattering coefficient $\beta=1$ and the pure-white atmospheric airlight $\alpha =1$). It is exciting that, although DehazeNet is optimized by the MSE loss function, it also achieves the best performance on the other types of evaluation matrices.
\begin{table*}
\center
\caption{The average results of MSE, SSIM, PSNR and WSNR on the synthetic images ($\beta=1$ and $\alpha=1$)}
\begin{tabular}{c|L{1.2cm}L{1.2cm}L{1.2cm}L{1.2cm}C{2.2cm}L{1.0cm}L{1.2cm}}
\hline
\hline
    Metric&ATM \cite{atm}&BCCR \cite{bccr}&FVR \cite{fvr}&DCP \cite{dcp}&CAP\footref{fn:cap}\cite{cap}&RF \cite{rf}&DehazeNet \tabularnewline \hline
    MSE&0.0689&0.0243&0.0155&0.0172& 0.0075 (0.0068)&\underline{\color{blue}{0.0070}}&\textbf{\color{red}{0.0062}} \tabularnewline \hline
    SSIM&0.9890&0.9963&0.9973&0.9981& \underline{\color{blue}{0.9991}} (0.9990)&0.9989&\textbf{\color{red}{0.9993}} \tabularnewline \hline
    PSNR&60.8612&65.2794&66.5450&66.7392&70.0029 (70.6581)&\underline{\color{blue}{70.0099}}&\textbf{\color{red}{70.9767}} \tabularnewline \hline
    WSNR&7.8492&12.6230&13.7236&13.8508&16.9873 (17.7839)&\underline{\color{blue}{17.1180}}&\textbf{\color{red}{18.0996}} \tabularnewline \hline
    \hline
\end{tabular}
\label{tab:synthetic_images}
\end{table*}

The dehazing effectiveness is sensitive to the haze density, and the performance with a different scattering coefficient $\beta$ could become much worse or better. Therefore, we propose an evaluation to analyze dehazing robustness to scattering coefficient $\beta\in\{0.75,1.0,1.25,1.5\}$, which is called as coefficient robustness evaluation (CRE). As shown in Table \ref{tab:synthetic_bs}, CAP \cite{cap} achieve better performances on the mist ($\beta=0.75$), but the dehazing performance reduces gradually when the amount of haze increases. The reason is that CAP estimates the medium transmission based on predicted scene depth and a assumed scattering coefficient ($\beta=1$). In \cite{rf}, 200 trees are used to build random forests for non-linear regression and shows greater coefficient robustness. However, the high-computation of random forests in every pixel constraints to its practicality. For DehazeNet, the medium transmission is estimated directly by a non-linear activation function (Maxout) in $F_1$, resulting in excellent robustness to the scattering coefficient.
\begin{table*}
\center
\caption{The MSE on the synthetic images by different scattering coefficient, image scale and atmospheric airlight}
\begin{tabular}{c|c|C{1.4cm}C{1.4cm}C{1.4cm}C{1.4cm}C{1.7cm}C{1.4cm}C{1.4cm}}
\hline
\hline
     \multicolumn{2}{c|}{Evaluation}&ATM \cite{atm}&BCCR \cite{bccr}&FVR \cite{fvr}&DCP \cite{dcp}&CAP\footref{fn:cap}\cite{cap}&RF \cite{rf}&DehazeNet \tabularnewline \hline
    \multirow{4}{*}{\minitab[c]{CRE\\$(\beta=)$}}&0.75&0.0581&0.0269&0.0122&0.0199&\textbf{\color{red}{0.0043}} (0.0042)&\underline{\color{blue}{0.0046}}&0.0063 \tabularnewline
    &1.00&0.0689&0.0243&0.0155&0.0172&0.0077 (0.0068)&\underline{\color{blue}{0.0070}}&\textbf{\color{red}{0.0062}} \tabularnewline
    &1.25&0.0703&0.0230&0.0219&0.0147&0.0141 (0.0121)&\underline{\color{blue}{0.0109}}&\textbf{\color{red}{0.0084}} \tabularnewline
    &1.50&0.0683&0.0219&0.0305&0.0134&0.0231 (0.0201)&\underline{\color{blue}{0.0152}}&\textbf{\color{red}{0.0127}} \tabularnewline \hline
    \multicolumn{2}{c|}{CRE Average}&0.0653&0.0254&0.0187&0.0177&0.0105 (0.0095)&\underline{\color{blue}{0.0094}}&\textbf{\color{red}{0.0084}} \tabularnewline
    \hline

    \multirow{4}{*}{\minitab[c]{ARE\\$(\alpha=)$}}&[1.0, 1.0, 1.0]&0.0689&0.0243&0.0155&0.0172&0.0075 (0.0068)&\underline{\color{blue}{0.0070}}&\textbf{\color{red}{0.0062}} \tabularnewline
    &[0.9, 1.0, 1.0]&0.0660&0.0266&0.0170&0.0210&0.0073 (0.0069)&\textbf{\color{red}{0.0071}}&\underline{\color{blue}{0.0072}} \tabularnewline
    &[1.0, 0.9, 1.0]&0.0870&0.0270&0.0159&0.0200&\textbf{\color{red}{0.0070}} (0.0067)&\underline{\color{blue}{0.0073}}&0.0074 \tabularnewline
    &[1.0, 1.0, 0.9]&0.0689&0.0239&0.0152&0.0186&\underline{\color{blue}{0.0081}} (0.0069)&0.0083&\textbf{\color{red}{0.0062}} \tabularnewline \hline
    \multicolumn{2}{c|}{ARE Average}&0.0727&0.0255&0.0159&0.0192&0.0075 (0.0068)&\underline{\color{blue}{0.0074}}&\textbf{\color{red}{0.0067}} \tabularnewline
    \hline

    \multirow{4}{*}{\minitab[c]{SRE\\$(s=)$}}&0.40&0.0450&0.0238&0.0155&0.0102&0.0137 (0.0084)&\underline{\color{blue}{0.0089}}&\textbf{\color{red}{0.0066}} \tabularnewline
    &0.60&0.0564&0.0223&0.0154&0.0137&0.0092 (0.0071)&\underline{\color{blue}{0.0076}}&\textbf{\color{red}{0.0060}} \tabularnewline
    &0.80&0.0619&0.0236&0.0155&0.0166&0.0086 (0.0066)&\underline{\color{blue}{0.0074}}&\textbf{\color{red}{0.0062}} \tabularnewline
    &1.00&0.0689&0.0243&0.0155&0.0172&0.0077 (0.0068)&\underline{\color{blue}{0.0070}}&\textbf{\color{red}{0.0062}} \tabularnewline \hline
    \multicolumn{2}{c|}{SRE Average}&0.0581&0.0235&0.0155&0.0144&0.0098 (0.0072)&\underline{\color{blue}{0.0077}}&\textbf{\color{red}{0.0062}} \tabularnewline
    \hline

    \multirow{5}{*}{\minitab[c]{NRE\\$(\sigma=)$}}
    &10&0.0541&0.0138&0.0150&0.0133&\underline{\color{blue}{0.0065}} (0.0070)&0.0086&\textbf{\color{red}{0.0059}} \tabularnewline
    &15&0.0439&0.0144&0.0148&0.0104&\underline{\color{blue}{0.0072}} (0.0074)&0.0112&\textbf{\color{red}{0.0061}} \tabularnewline
    &20&--&0.0181&0.0151&0.0093&\underline{\color{blue}{0.0083}} (0.0085)&0.0143&\textbf{\color{red}{0.0058}} \tabularnewline
    &25&--&0.0224&0.0150&\underline{\color{blue}{0.0082}}&0.0100 (0.0092)&0.0155&\textbf{\color{red}{0.0051}} \tabularnewline
    &30&--&0.0192&0.0151&\underline{\color{blue}{0.0085}}&0.0119 (0.0112)&0.0191&\textbf{\color{red}{0.0049}} \tabularnewline
    \hline
    \multicolumn{2}{c|}{NRE Average}
    &--&0.0255&0.0150&0.0100&\underline{\color{blue}{0.0088}} (0.0087)&0.0137&\textbf{\color{red}{0.0055}} \tabularnewline

    \hline
    \hline
    \end{tabular}
\label{tab:synthetic_bs}
\end{table*}

\begin{figure}[!t]
\centering
\subfigure{
\includegraphics[width=0.3\linewidth]{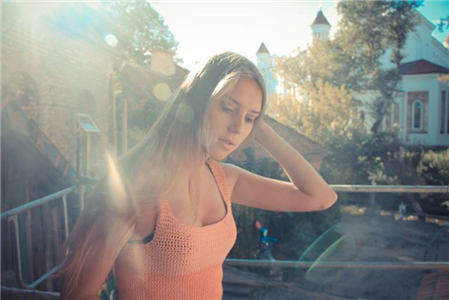}
\includegraphics[width=0.3\linewidth]{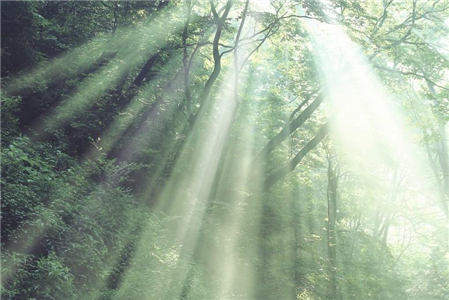}
\includegraphics[width=0.3\linewidth]{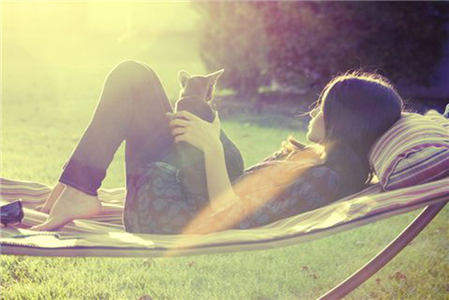}}
\subfigure{
\includegraphics[width=0.3\linewidth]{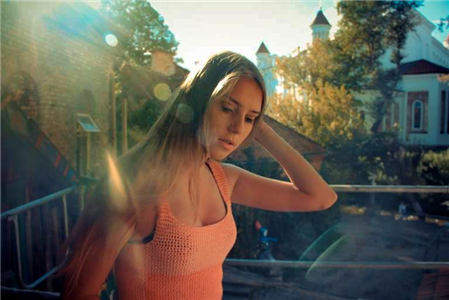}
\includegraphics[width=0.3\linewidth]{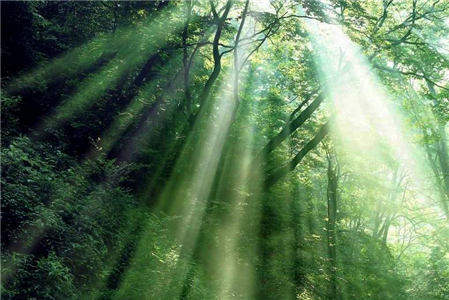}
\includegraphics[width=0.3\linewidth]{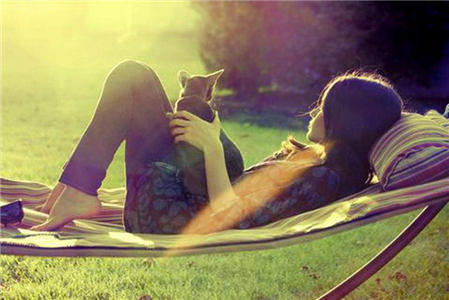}}
\caption{Image enhancement for anti-halation by DehazeNet}
\label{fig:halation}
\end{figure}

Due to the color offset of haze particles and light sources, the atmosphere airlight is not a proper pure-white. An airlight robustness evaluation (ARE) is proposed to analyze the dehazing methods for different atmosphere airlight $\alpha$. Although DehazeNet is trained from the samples generated by setting $\alpha=1$, it also achieves the greater robustness on the other values of atmosphere airlight. In particular, DehazeNet performs better than the other methods when sunlight haze is $[1.0, 1.0, 0.9]$. Therefore, DehazeNet could also be applied to remove halation, which is a bright ring surrounding a source of light as shown in Fig. \ref{fig:halation}.

The view field transformation and image zoom occur often in real-world applications. The scale robustness evaluation (SRE) is used to analyze the influence from the scale variation. Compared with the same state-of-the-art methods in OPE, there are 4 scale coefficients $s$ selected from 0.4 to 1.0 to generate different scale images for SER. In Table \ref{tab:synthetic_bs}, DehazeNet shows excellent robustness to the scale variation due to the multi-scale mapping in $F_2$. The single scale used in CAP \cite{cap}, DCP \cite{dcp} and ATM \cite{atm} results in a different prediction accuracy on a different scale. When an image shrinks, an excessively large-scale processing neighborhood will lose the image's details. Therefore, the multi-scale mapping in DehazeNet provides a variety of filters to merge multi-scale features, and it achieves the best scores under all of different scales.

In most situations, noise is random produced by the sensor or camera circuitry, which will bring in estimation error. We also discuss the influences of varying degrees of image noise to our method. As a basic noise model, additive white Gaussian (AWG) noise with standard deviation  $\sigma\in\{10,15,20,25\}$ is used for noise robustness evaluation (NRE). Benefiting from the Maxout suppression in $F_1$  and the local extremum in $F_3$, DehazeNet performs more robustly in NRE than the others do. RF \cite{rf} has a good performance in most of the evaluations but fails in NRE, because the feature values of patches are sorted to break the correlation between the medium transmission and the image content, which will also magnify the effect of outlier.

\subsection{Qualitative results on real-world images}

\begin{figure*}[!t]
\centering
\includegraphics[height=1.3cm]{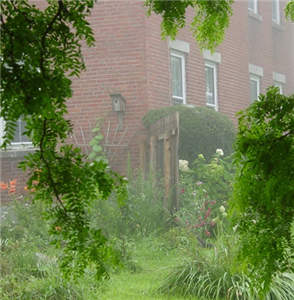}
\includegraphics[height=1.3cm]{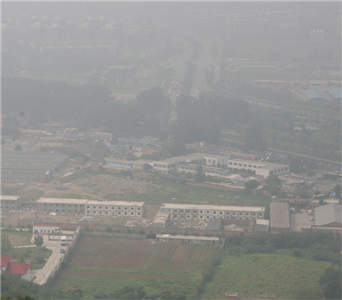}
\includegraphics[height=1.3cm]{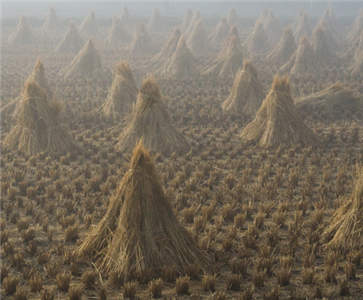}
\includegraphics[height=1.3cm]{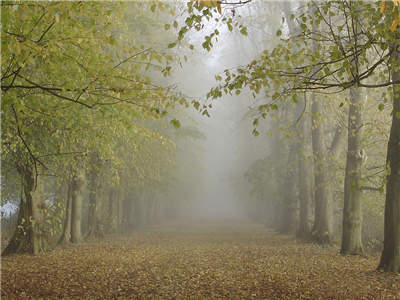}
\includegraphics[height=1.3cm]{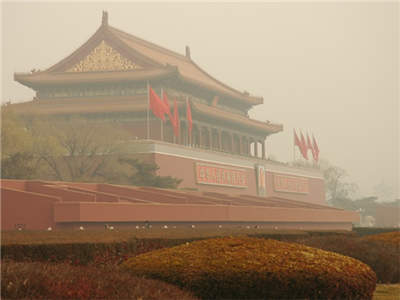}
\includegraphics[height=1.3cm]{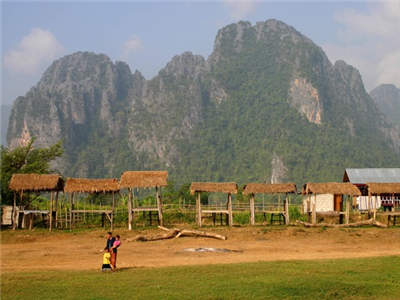}
\includegraphics[height=1.3cm]{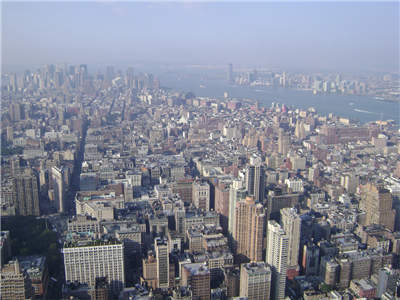}
\includegraphics[height=1.3cm]{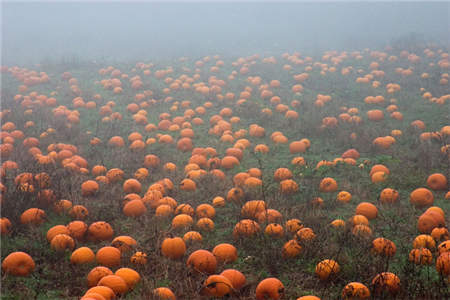}
\includegraphics[height=1.3cm]{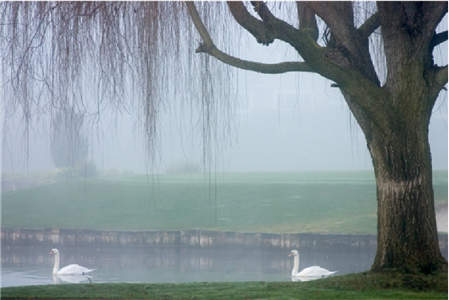}
\includegraphics[height=1.3cm]{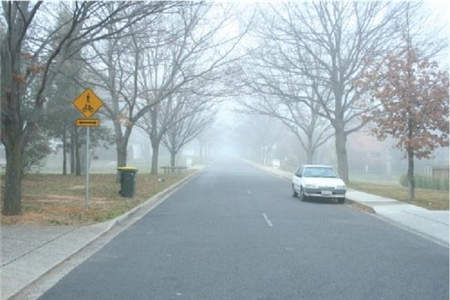}
\\[0.08cm]
\includegraphics[height=1.3cm]{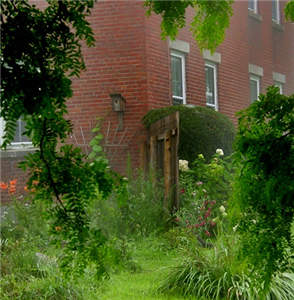}
\includegraphics[height=1.3cm]{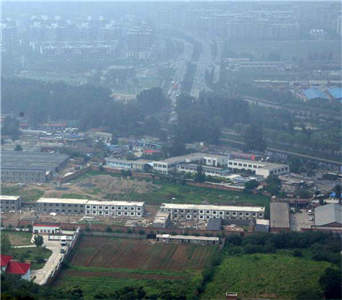}
\includegraphics[height=1.3cm]{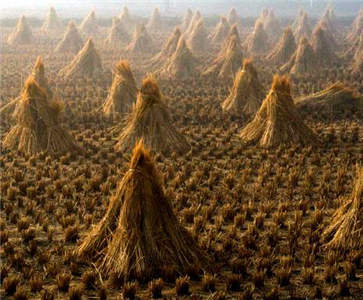}
\includegraphics[height=1.3cm]{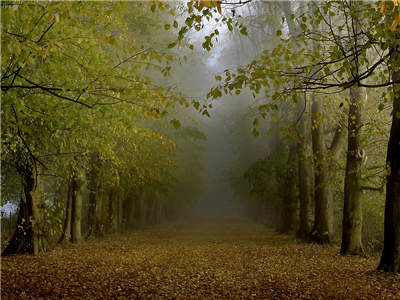}
\includegraphics[height=1.3cm]{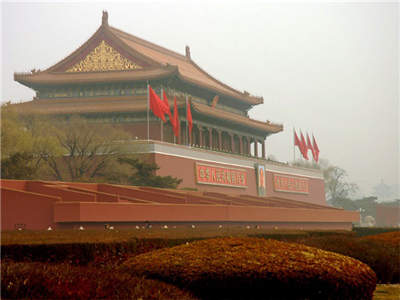}
\includegraphics[height=1.3cm]{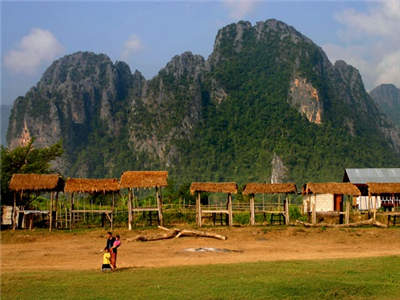}
\includegraphics[height=1.3cm]{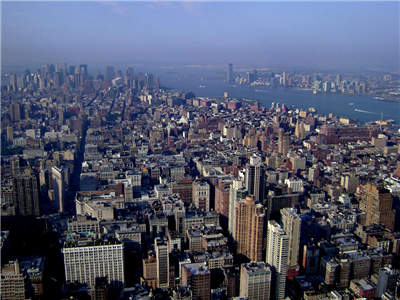}
\includegraphics[height=1.3cm]{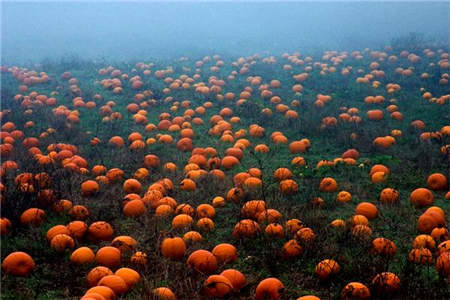}
\includegraphics[height=1.3cm]{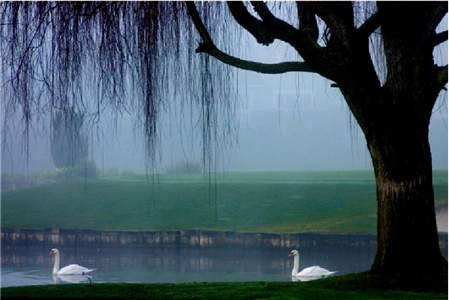}
\includegraphics[height=1.3cm]{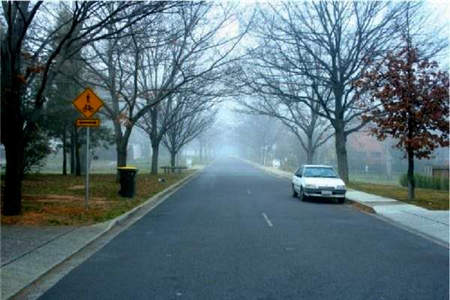}
\\[0.08cm]
\includegraphics[height=1.3cm]{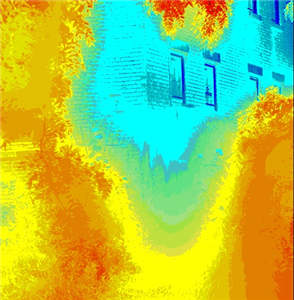}
\includegraphics[height=1.3cm]{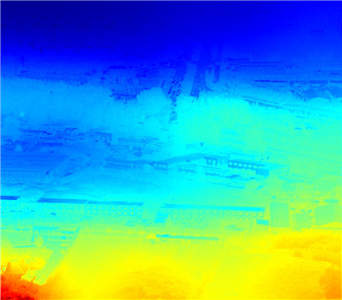}
\includegraphics[height=1.3cm]{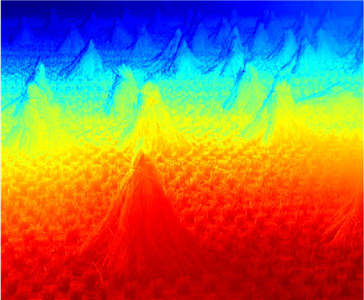}
\includegraphics[height=1.3cm]{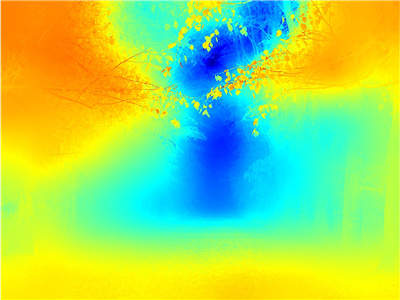}
\includegraphics[height=1.3cm]{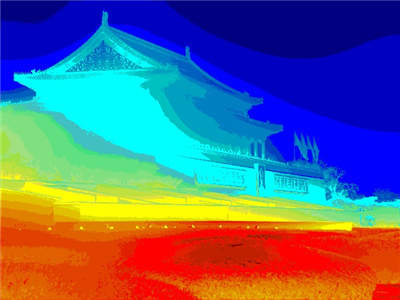}
\includegraphics[height=1.3cm]{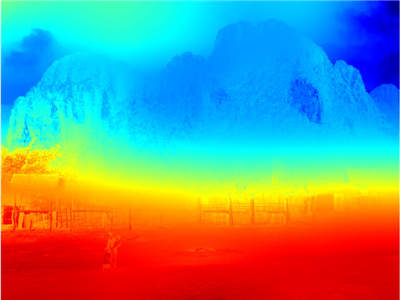}
\includegraphics[height=1.3cm]{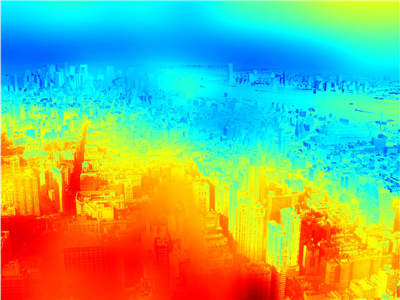}
\includegraphics[height=1.3cm]{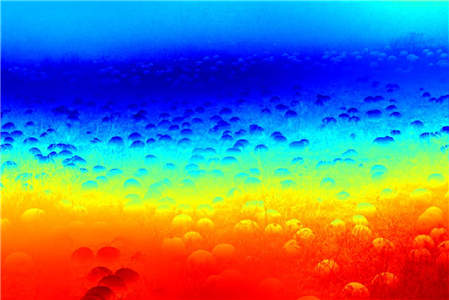}
\includegraphics[height=1.3cm]{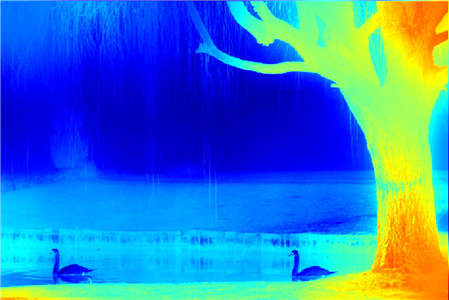}
\includegraphics[height=1.3cm]{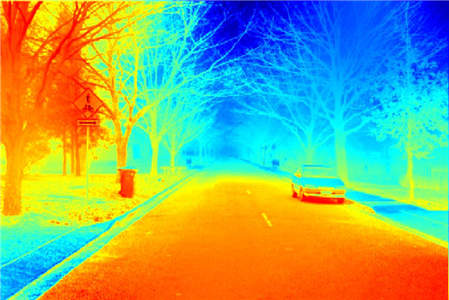}
\caption{The haze-free images and depth maps restored by DehazeNet}
\label{fig:more_result}
\end{figure*}

\begin{figure*}[!t]
\centering
\subfigure{
\includegraphics[width=0.12\linewidth]{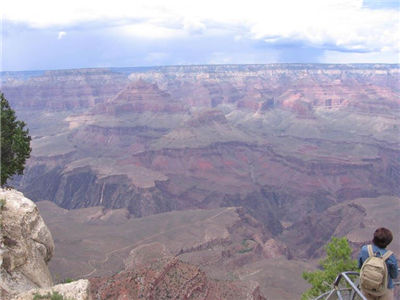}
\includegraphics[width=0.12\linewidth]{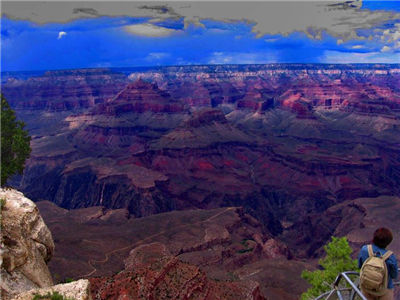}
\includegraphics[width=0.12\linewidth]{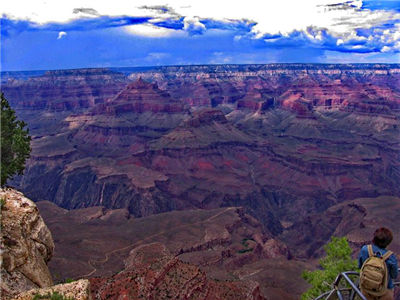}
\includegraphics[width=0.12\linewidth]{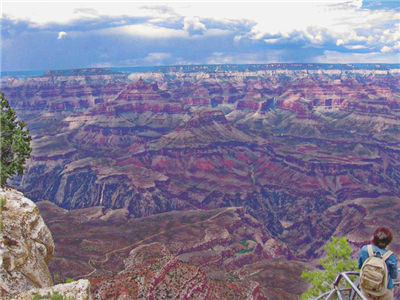}
\includegraphics[width=0.12\linewidth]{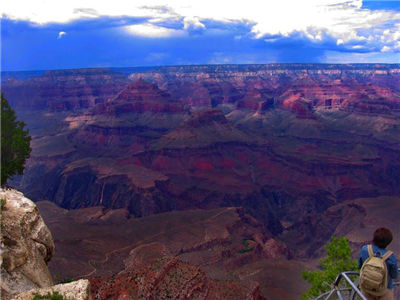}
\includegraphics[width=0.12\linewidth]{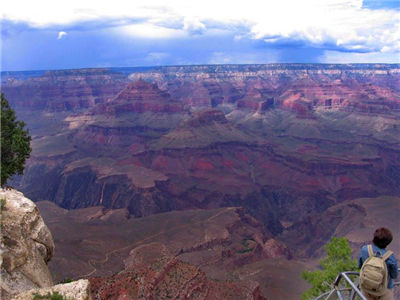}
\includegraphics[width=0.12\linewidth]{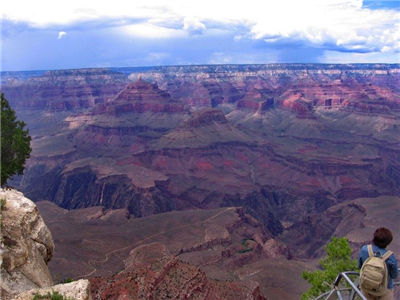}
\includegraphics[width=0.12\linewidth]{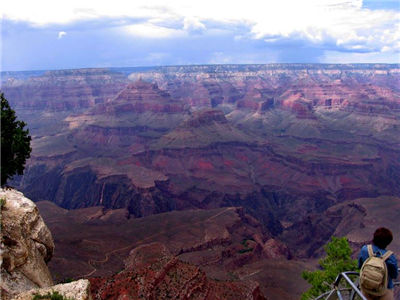}
}
\subfigure{
\includegraphics[width=0.12\linewidth]{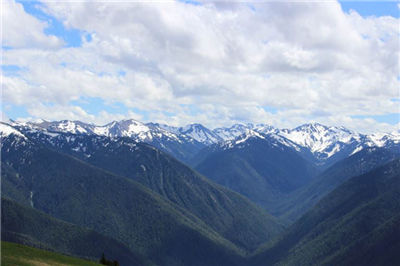}
\includegraphics[width=0.12\linewidth]{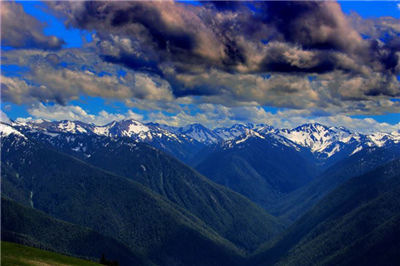}
\includegraphics[width=0.12\linewidth]{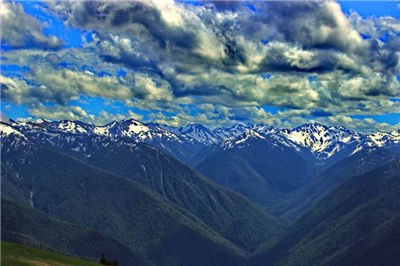}
\includegraphics[width=0.12\linewidth]{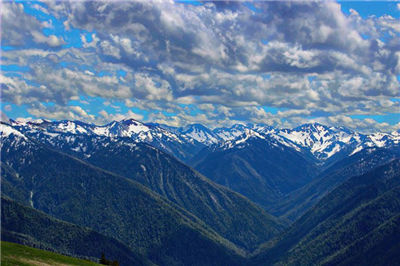}
\includegraphics[width=0.12\linewidth]{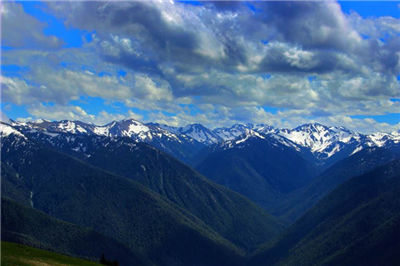}
\includegraphics[width=0.12\linewidth]{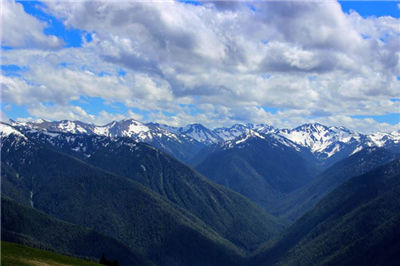}
\includegraphics[width=0.12\linewidth]{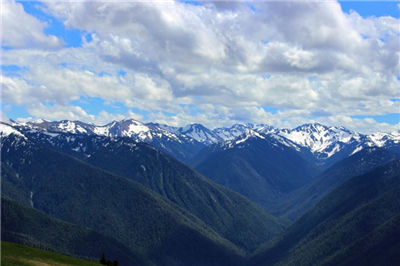}
\includegraphics[width=0.12\linewidth]{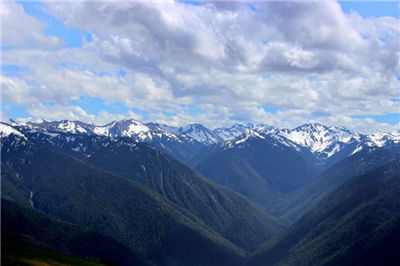}
}
\subfigure{
\includegraphics[width=0.12\linewidth]{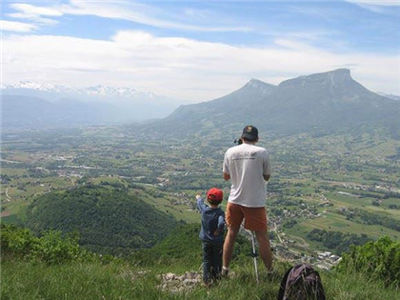}
\includegraphics[width=0.12\linewidth]{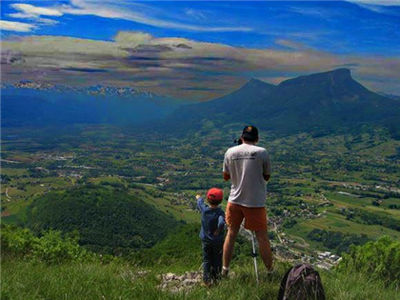}
\includegraphics[width=0.12\linewidth]{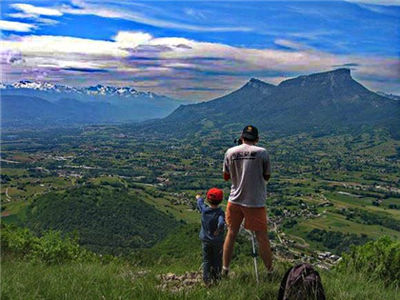}
\includegraphics[width=0.12\linewidth]{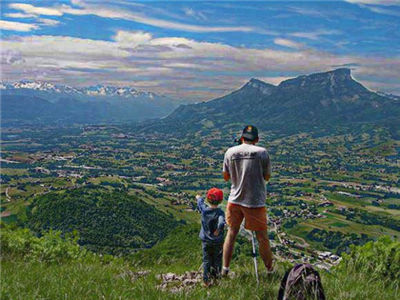}
\includegraphics[width=0.12\linewidth]{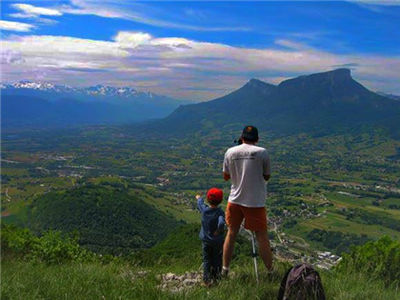}
\includegraphics[width=0.12\linewidth]{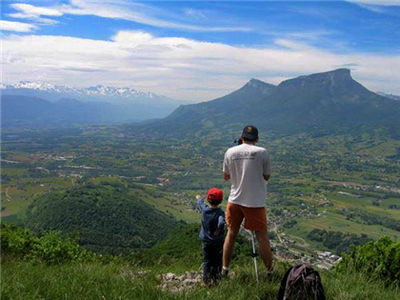}
\includegraphics[width=0.12\linewidth]{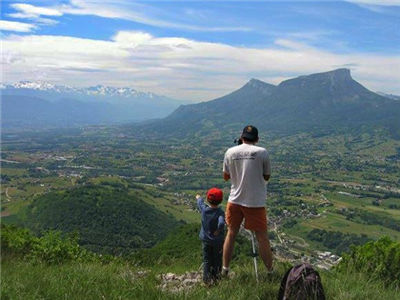}
\includegraphics[width=0.12\linewidth]{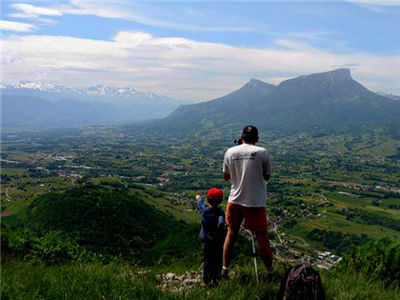}
}
\subfigure{
\includegraphics[width=0.12\linewidth]{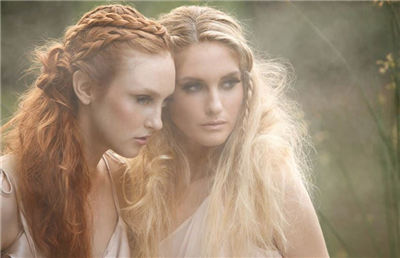}
\includegraphics[width=0.12\linewidth]{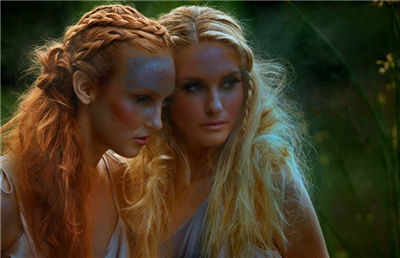}
\includegraphics[width=0.12\linewidth]{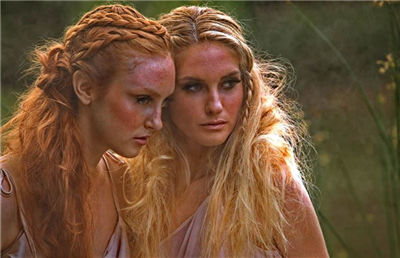}
\includegraphics[width=0.12\linewidth]{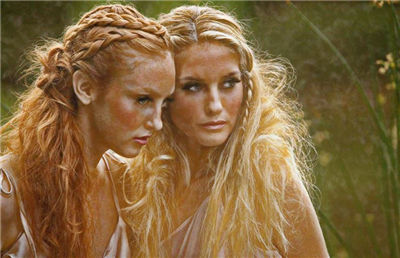}
\includegraphics[width=0.12\linewidth]{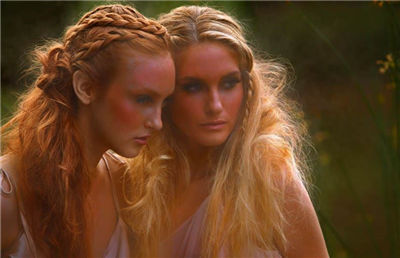}
\includegraphics[width=0.12\linewidth]{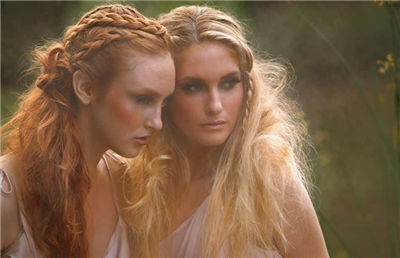}
\includegraphics[width=0.12\linewidth]{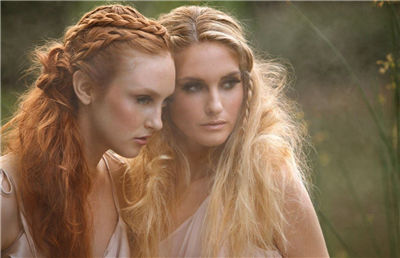}
\includegraphics[width=0.12\linewidth]{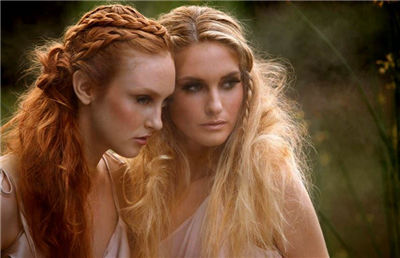}
}
\subfigure{
\includegraphics[width=0.12\linewidth]{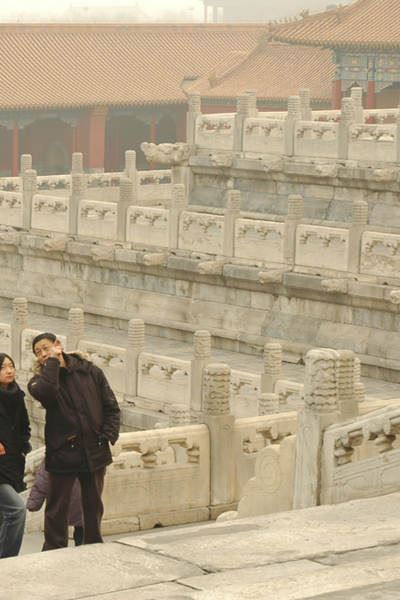}
\includegraphics[width=0.12\linewidth]{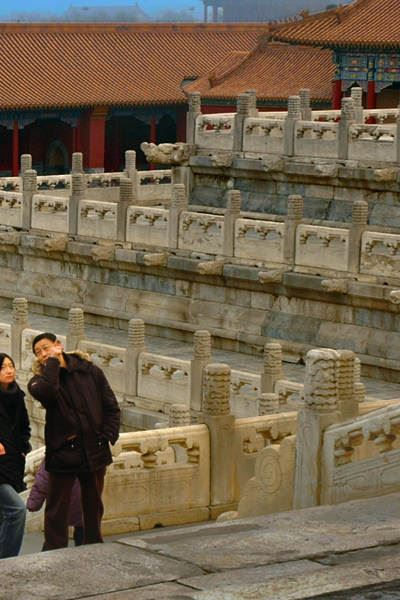}
\includegraphics[width=0.12\linewidth]{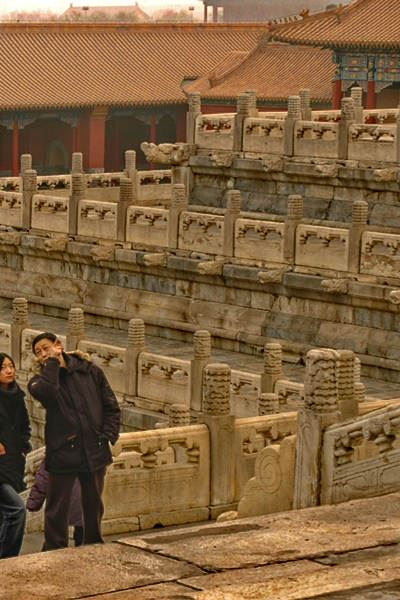}
\includegraphics[width=0.12\linewidth]{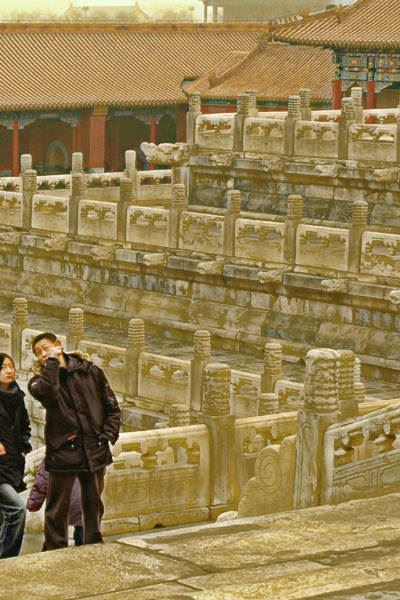}
\includegraphics[width=0.12\linewidth]{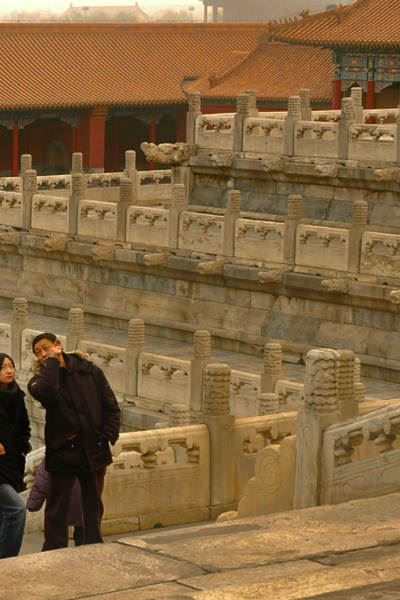}
\includegraphics[width=0.12\linewidth]{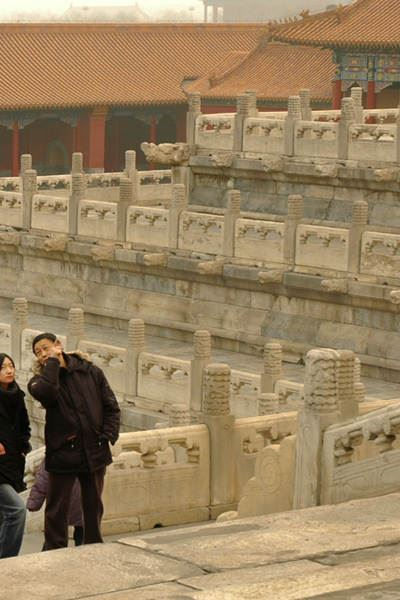}
\includegraphics[width=0.12\linewidth]{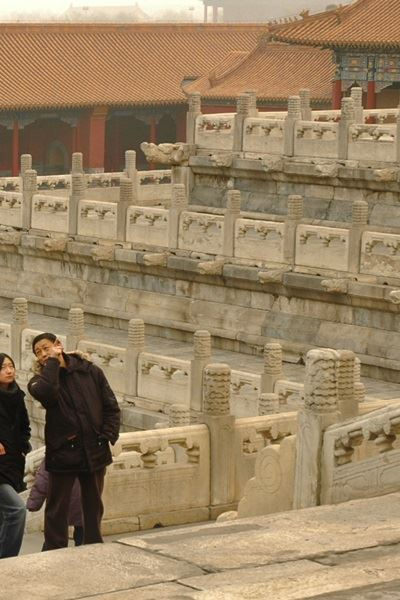}
\includegraphics[width=0.12\linewidth]{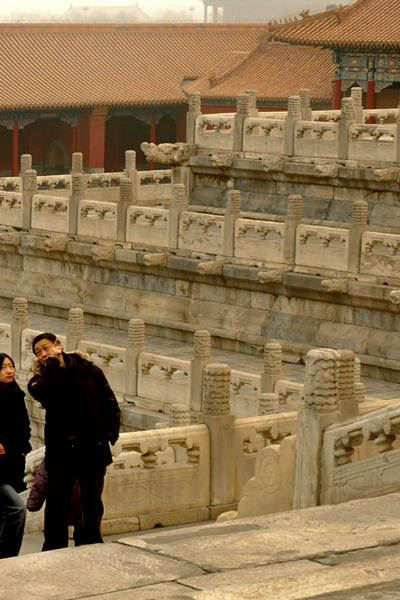}
}
\footnotesize  \begin{tabular}{C{1.925cm}C{1.925cm}C{1.925cm}C{1.925cm}C{1.925cm}C{1.925cm}C{1.925cm}C{1.925cm}}
(a) Hazy image&(b) ATM \cite{atm}&(c) BCCR \cite{bccr}&(d) FVR \cite{fvr}&(e) DCP \cite{dcp}&(f) CAP \cite{cap}&(g) RF \cite{rf}&(h) DehazeNet
\end{tabular}
\caption{Qualitative comparison of different methods on real-world images.}
\label{fig:realworld}
\end{figure*}

Fig. \ref{fig:more_result} shows the dehazing results and depth maps restored by DehazeNet, and more results and comparisons can be found at \url{http://caibolun.github.io/DehazeNet/}. Because all of the dehazing algorithms can obtain truly good results on general outdoor images, it is difficult to rank them visually. To compare them, this paper focuses on 5 identified challenging images in related studies \cite{dcp,rf,cap}. These images have large white or gray regions that are hard to handle, because most existing dehazing algorithms are sensitive to the white color. Fig. \ref{fig:realworld} shows a qualitative comparison with six state-of-the-art dehazing algorithms on the challenging images. Fig. \ref{fig:realworld} (a) depicts the hazy images to be dehazed, and Fig. \ref{fig:realworld} (b-g) shows the results of ATM \cite{atm}, BCCR \cite{bccr}, FVR \cite{fvr}, DCP \cite{dcp}, CAP \cite{cap} and RF \cite{rf}, respectively. The results of DehazeNet are given in Fig. \ref{fig:realworld} (h).

The sky region in a hazy image is a challenge of dehazing, because clouds and haze are similar natural phenomenons with the same atmospheric scattering model. As shown in the first three figures, most of the haze is removed in the (b-d) results, and the details of the scenes and objects are well restored. However, the results significantly suffer from over-enhancement in the sky region. Overall, the sky region of these images is much darker than it should be or is oversaturated and distorted. Haze generally exists only in the atmospheric surface layer, and thus the sky region almost does not require handling. Based on the learning framework, CAP and RF avoid color distortion in the sky, but non-sky regions are enhanced poorly because of the non-content regression model (for example, the rock-soil of the first image and the green flatlands in the third image). DehazeNet appears to be capable of finding the sky region to keep the color, and assures a good dehazing effect in other regions. The reason is that the patch attribute can be learned in the hidden layer of DehazeNet, and it contributes to the dehazing effects in the sky.

Because transmission estimation based on priors are a type of statistics, which might not work for certain images. The fourth and fifth figures are determined to be failure cases in \cite{dcp}. When the scene objects are inherently similar to the atmospheric light (such as the fair-skinned complexion in the fourth figure and the white marble in the fifth figure), the estimated transmission based on priors (DCP, BCCR, FVR) is not reliable. Because the dark channel has bright values near such objects, and FVR and BCCR are based on DCP which has an inherent problem of overestimating the transmission. CAP and RF learned from a regression model is free from oversaturation, but underestimates the haze degree in the distance (see the brown hair in the fourth image and the red pillar in the fifth image). Compared with the six algorithms, our results avoid image oversaturation and retain the dehazing validity due to the non-linear regression of DehazeNet.

\section{Conclusion}\label{sec:conclusion}

In this paper, we have presented a novel deep learning approach for single image dehazing. Inspired by traditional haze-relevant features and dehazing methods, we show that medium transmission estimation can be reformulated into a trainable end-to-end system with special design, where the feature extraction layer and the non-linear regression layer are distinguished from classical CNNs. In the first layer $F_1$, Maxout unit is proved similar to the priori methods, and it is more effective to learn haze-relevant features. In the last layer $F_4$, a novel activation function called BReLU is instead of ReLU or Sigmoid to keep bilateral restraint and local linearity for image restoration. With this lightweight architecture, DehazeNet achieves dramatically high efficiency and outstanding dehazing effects than the state-of-the-art methods.

Although we successfully applied a CNN for haze removal, there are still some extensibility researches to be carried out. That is, the atmospheric light $\alpha$ cannot be regarded as a global constant, which will be learned together with medium transmission in a unified network. Moreover, we think atmospheric scattering model can also be learned in a deeper neural network, in which an end-to-end mapping between haze and haze-free images can be optimized directly without the medium transmission estimation.
We leave this problem for future research.


\ifCLASSOPTIONcaptionsoff
  \newpage
\fi

\bibliographystyle{IEEEtran}
\bibliography{refs}




\end{document}